\pdfoutput=1

\documentclass[11pt]{article}

\usepackage[]{EMNLP2023}

\usepackage{times}
\usepackage{latexsym}

\usepackage[T1]{fontenc}

\usepackage[utf8]{inputenc}

\usepackage{microtype}

\usepackage{inconsolata}
\usepackage{graphicx}
\usepackage{times}
\usepackage{latexsym}
\usepackage{multirow}
\usepackage{pifont}
\usepackage{booktabs}
\usepackage[T2A,T1]{fontenc}

\usepackage{CJKutf8}  
\usepackage{float}  
\usepackage{tabularx}
\usepackage{longtable}
\usepackage[utf8]{inputenc}
\usepackage{CJKutf8}   
\usepackage{tcolorbox}  
\usepackage{setspace} 
\usepackage[russian,english]{babel}
\usepackage{dblfloatfix}  
\usepackage{graphicx}
\usepackage{subcaption}
\usepackage{placeins} 
\usepackage{lipsum}       
\usepackage{capt-of}      
\usepackage{cuted}

\title{MAATS: A Multi-Agent Automated Translation System Based on MQM Evaluation}

\begin{document}
\maketitle
\begin{abstract}
We present \textbf{MAATS}, a \textbf{M}ulti \textbf{A}gent \textbf{A}utomated \textbf{T}ranslation \textbf{S}ystem that leverages the Multidimensional Quality Metrics (MQM) framework as a fine-grained signal for error detection and refinement. MAATS employs multiple specialized AI agents, each focused on a distinct MQM category (e.g., Accuracy, Fluency, Style, Terminology), followed by a synthesis agent that integrates the annotations to iteratively refine translations. This design contrasts with conventional single-agent methods that rely on self-correction. 

Evaluated across diverse language pairs and Large Language Models (LLMs), MAATS outperforms zero-shot and single-agent baselines with statistically significant gains in both automatic metrics and human assessments. It excels particularly in semantic accuracy, locale adaptation, and linguistically distant language pairs. Qualitative analysis highlights its strengths in multi-layered error diagnosis, omission detection across perspectives, and context-aware refinement. By aligning modular agent roles with interpretable MQM dimensions, MAATS narrows the gap between black-box LLMs and human translation workflows, shifting focus from surface fluency to deeper semantic and contextual fidelity. Code and data:

\href{https://github.com/maats0519/maats_mqm.git}{\texttt{https://github.com/maats0519/maats\_mqm}}.

\end{abstract}

\section{Introduction}
\begin{figure}[t] 
    \centering
    \includegraphics[width=1\linewidth]{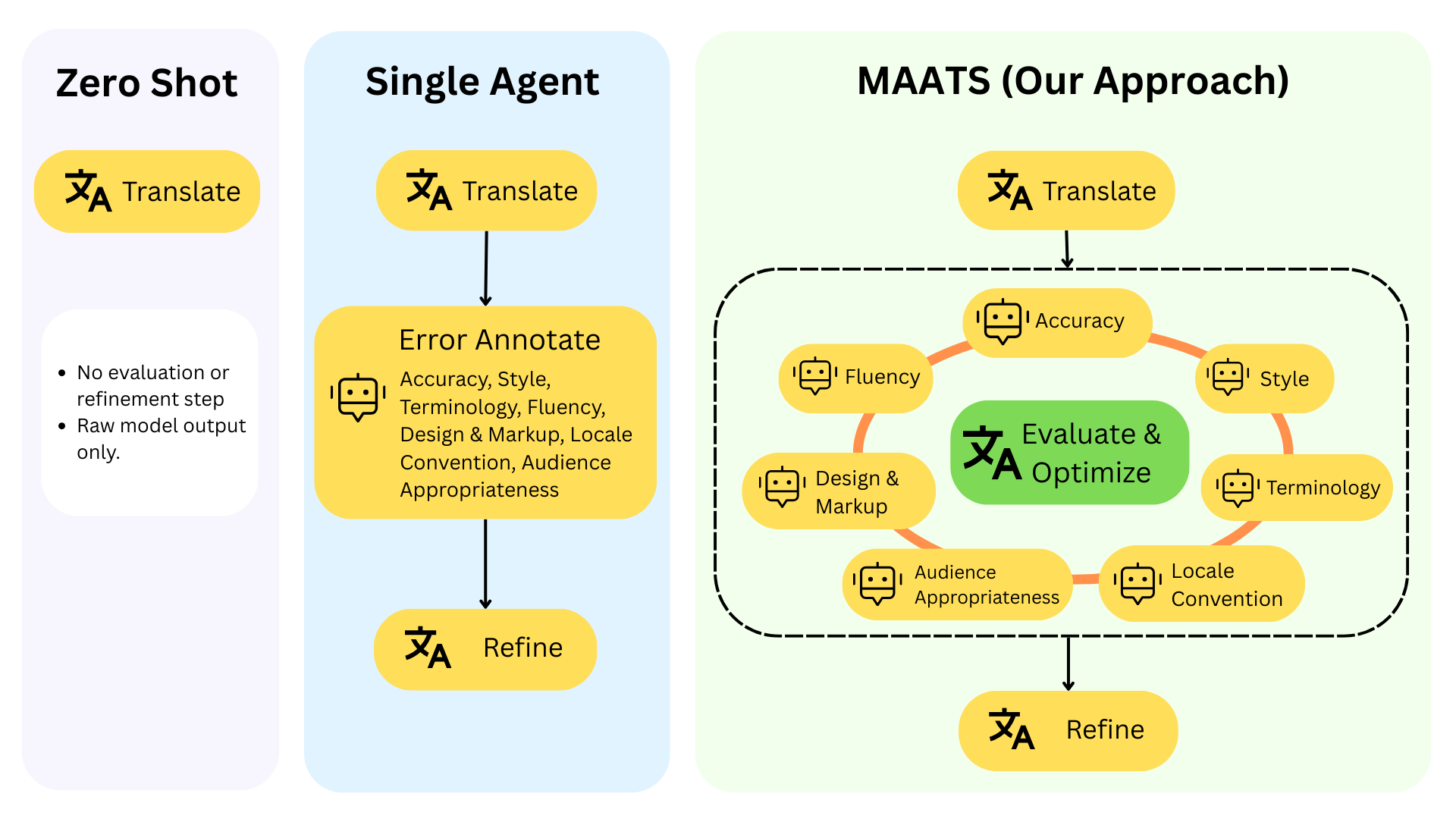}
    \vspace{-0.8em} 
    \caption{Comparison of Zeor-Shot, Single-Agent and MAATS Pipelines. MAATS assigns MQM dimensions to specialized agents, whose outputs are synthesized by a central agent \cite{lommel2024multirangetheorytranslationquality}.}
    \label{fig:maats_pipeline}
    \vspace{-1.5em} 
\end{figure}

\begin{figure*}[htbp!]
    \centering
    \includegraphics[width=\linewidth]{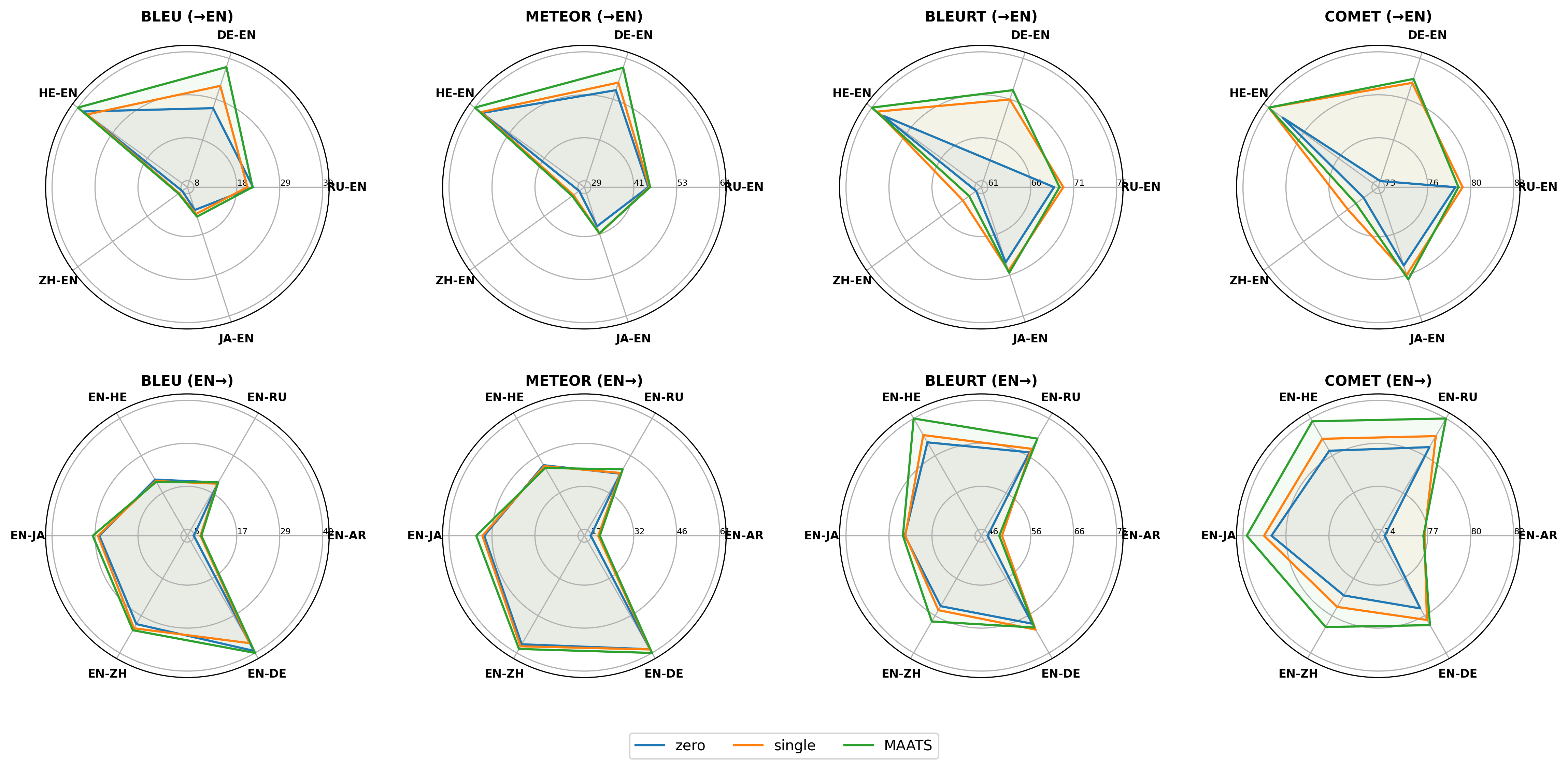}
    \caption{Translation Metrics Comparison for GPT-4o Across Approaches and Language Pairs. } Full-scale versions see Table \ref{fig:full scale}.
    \label{fig:fig-gpt4o}
\end{figure*}

While both multi-agent systems and machine translation have seen rapid progress \cite{manakhimova-etal-2023-linguistically,peng2023makingchatgptmachinetranslation,jiao2023chatgptgoodtranslatoryes}, the use of explicitly specialized agents assigned to distinct evaluation tasks remains underexplored. Despite advances that enable LLMs to rival dedicated MT systems, even top-performing models still produce subtle mistranslations, omissions, stylistic errors, and terminology inconsistencies \cite{Freitag_2021,jiao2023chatgptgoodtranslatoryes,yan2024benchmarkinggpt4humantranslators}.

Researchers are drawing inspiration from professional human translation workflows, which typically involve a multi-stage process with collaboration between translators and editors \cite{yan-etal-2014-two,10.1007/978-3-031-60615-1_8,wu2024perhapshumantranslationharnessing}. This insight has led to refinement approaches for LLMs, including single-agent methods where a model attempts to critique and correct its own output\cite{feng2024tearimprovingllmbasedmachine}. Although such self-refinement can reduce errors, a single model can struggle to identify all issues due to potential biases or a lack of diverse expertise \cite{xu2024prideprejudicellmamplifies,kamoi2024evaluatingllmsdetectingerrors}, which may have implications on errors in machine translation.

This paper introduces MAATS (Multi-Agent Automated Translation System), a novel framework designed explicitly to model a collaborative annotation + refinement process using specialized LLM-based agents. It uses the MQM framework as a shared language for annotating translation errors across multiple dimensions like accuracy, fluency, and style\cite{lommel2024multirangetheorytranslationquality}. All annotations are synthesized by an Editor agent to produce a refined translation.

Our findings are: 1) MAATS consistently outperforms two baselines: zero-shot and single-agent approaches by correcting more critical errors with higher scores on both standard and neural metrics 2) Statistical analysis confirms that MAATS shows significant neural metric gains for linguistically distant pairs and weaker base LLMs. 3) Compared to human MQM annotations, MAATS increases true positives and reduces false negatives over both baselines. 4) Human evaluation shows professional translators ranked MAATS highest over two baselines across models. 5) Qualitative analysis with case studies shows MAATS performs deep error analysis and context-aware translation refinement.

\vspace{-0.3 em}
\section{Method}
\subsection{MAATS Design}

Unlike traditional approaches that combine translation, evaluation, and refinement in one linear step \cite{feng2024tearimprovingllmbasedmachine}, MAATS introduces a modular framework where specialized LLM agents collaboratively enhance translation quality through targeted error detection and correction.

Shown in Figure 1, the process begins with a Translator Agent, which generates the initial translation using base LLMs. This output is then evaluated by a set of MQM Evaluator Agents, each aligned with a specific MQM category: Accuracy, Fluency, Locale Convention, Audience Appropriateness, Style, Terminology, and Design \& Markup\cite{lommel2024multirangetheorytranslationquality}. These agents independently annotate translation errors within their domain with critical, major, minor severity levels. All annotations are passed to a centralized Editor Agent, which synthesizes and prioritizes the suggested corrections. The Editor prioritizes resolving critical issues, then addresses less severe ones by severity. This modular design avoids iterative feedback loops and ensures efficient integration of evaluations without redundancy or annotation conflicts. Overall, the MAATS architecture delivers higher translation quality and interpretability by simulating the collaborative workflows of human translator-editor teams.

\subsection{Prompts and Models}

MAATS employs distinct prompts for the Translator, specialized MQM Evaluators (See Appendix~\ref{AccuracyPrompt}), and Editor agents (Appendix~\ref{MAATS editor prompt}). These prompts were designed using a few-shot approach based on real examples adapted from Unbabel’s Typology 3.0.\footnote{\url{https://help.unbabel.com/hc/en-us/articles/6444304419479-Annotation-Guidelines-Typology-3-0}} For the Single-Agent baseline, a self-refinement prompt is used (Appendix~\ref{single-agent prompt}). Experiments were conducted using three state-of-the-art large LLMs as base models: Claude-3-haiku, Gemini-2.0-flash, and GPT-4o \cite{anthropic2024claude3,google2025gemini,openai2024gpt4o}. These models served as the base for all three approaches: zero-shot, single-agent, and MAATS. Implementation details for both MAATS and the Single-Agent baseline are provided in Appendix~\ref{appendix:A}.

\begin{figure}[H]
    \centering
    \includegraphics[width=1 \linewidth]{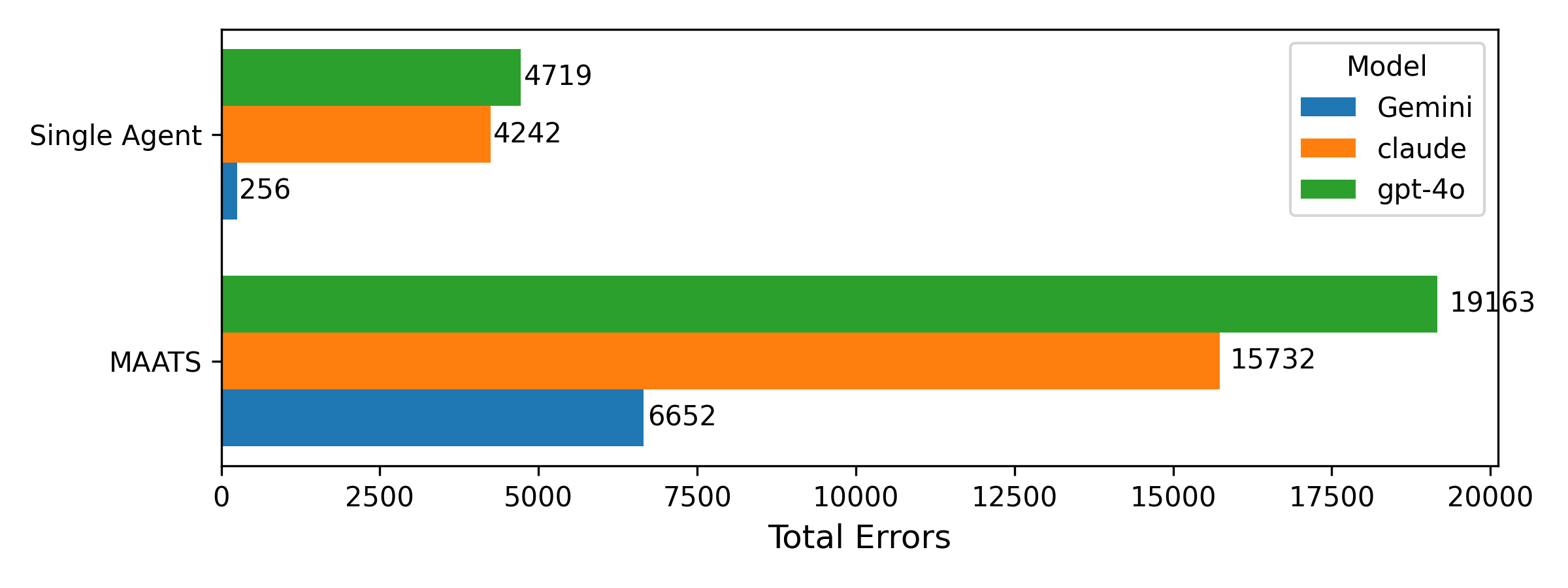}

    \caption{MAATS vs. Single-Agent Total Annotation Counts. GPT found the most errors, followed by Claude and Gemini
}
    \label{fig:overall maats vs single}
\end{figure}
\vspace{-1.5 em}
\subsection{Tasks and Evaluation}

To evaluate the MAATS system, translation tasks were conducted bidirectionally between English

\noindent and six target languages: German (DE), Hebrew (HE), Japanese (JA), Russian (RU), Chinese (ZH), and Arabic (AR) from WMT 2023\footnote{WMT 2023: \url{https://www.statmt.org/wmt23/}} and WMT 2024\footnote{WMT 2024: \url{https://www.statmt.org/wmt24/}}.

To comprehensively evaluate the effectiveness of the MAATS system, we conducted six distinct evaluations, each addressing a different aspect of translation quality. \\ \textbf{Automated Metric Comparison.} MAATS was benchmarked against two baselines using standard evaluation metrics, including BLEU, METEOR, BLEURT, and  COMET \cite{papineni-etal-2002-bleu, banerjee-lavie-2005-meteor, sellam-etal-2020-bleurt, rei-etal-2020-comet}. \\ \textbf{Statistical Testing.} ANOVA and paired bootstrap resampling were applied to assess the reliability of observed improvements, confirming that MAATS's gains were statistically significant. \\ \textbf{Human-Annotated MQM Evaluation.} We validated MAATS's error detection performance by comparing its outputs against human-labeled MQM reference data using confusion matrices \cite{freitag2021experts}. \\ \textbf{Human Preference Ranking.} Professional bilingual translators independently ranked translations from MAATS, single-agent, and zero-shot systems. \\ \textbf{Qualitative Case Studies.} Three detailed case studies highlighted MAATS's strengths in handling cultural references, multi-perspective error detection, and context-aware refinement.

\section{Results}

\noindent\textbf{MAATS vs. Baselines Annotation.} We evaluated the scalability and sensitivity in error annotation of MAATS across different LLMs. MAATS identified 41,547 translation issues across all language pairs, 
\begin{figure}[H]
    \centering
    \includegraphics[width=0.9\linewidth]{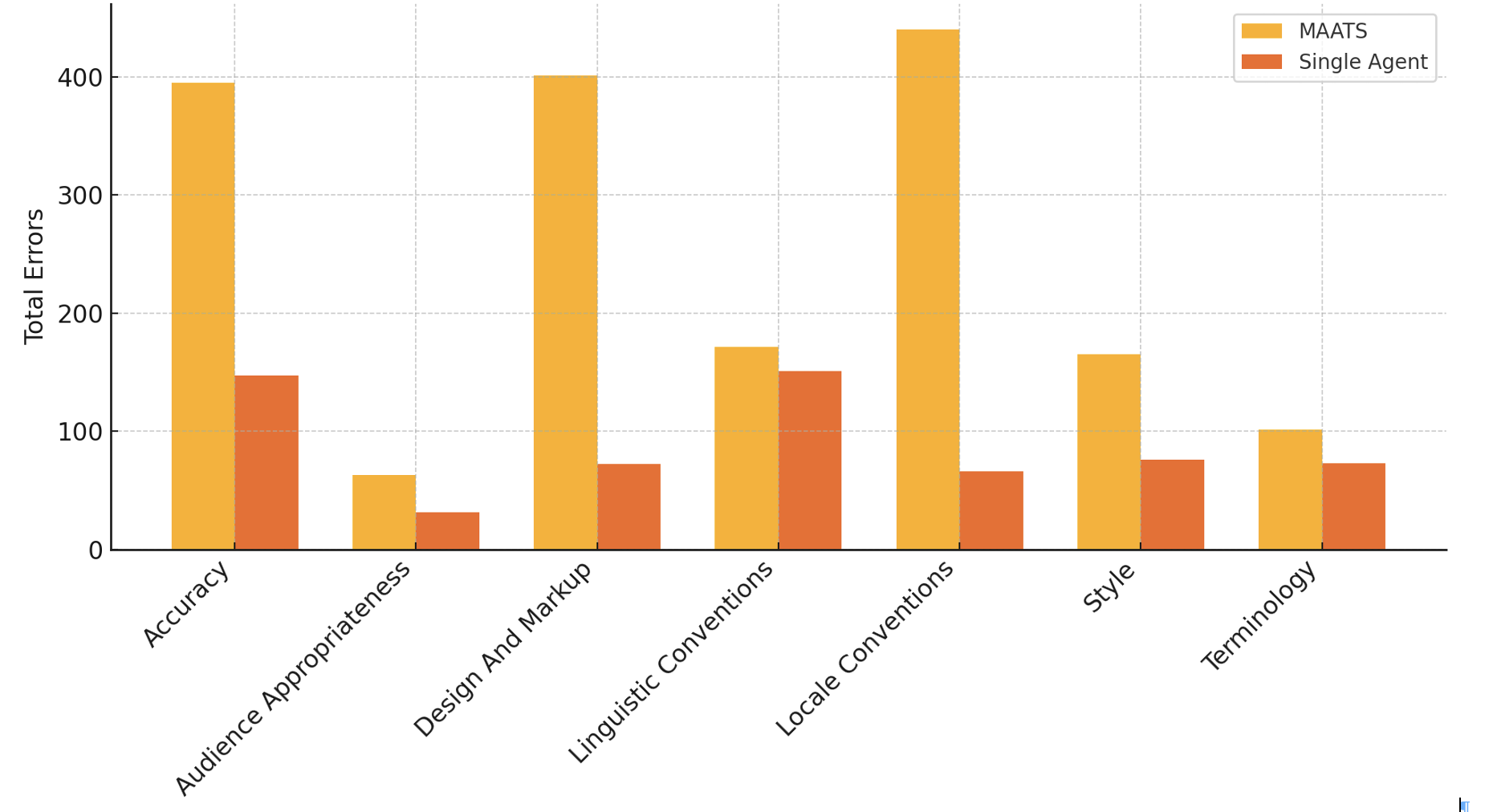}

    \caption{Chinese to English Annotation Analysis Using MAATS and Single Agent }
    \label{fig:zh_en maats vs single}
\end{figure}

\begin{figure}[H]
    \centering
    \includegraphics[width=0.9\linewidth]{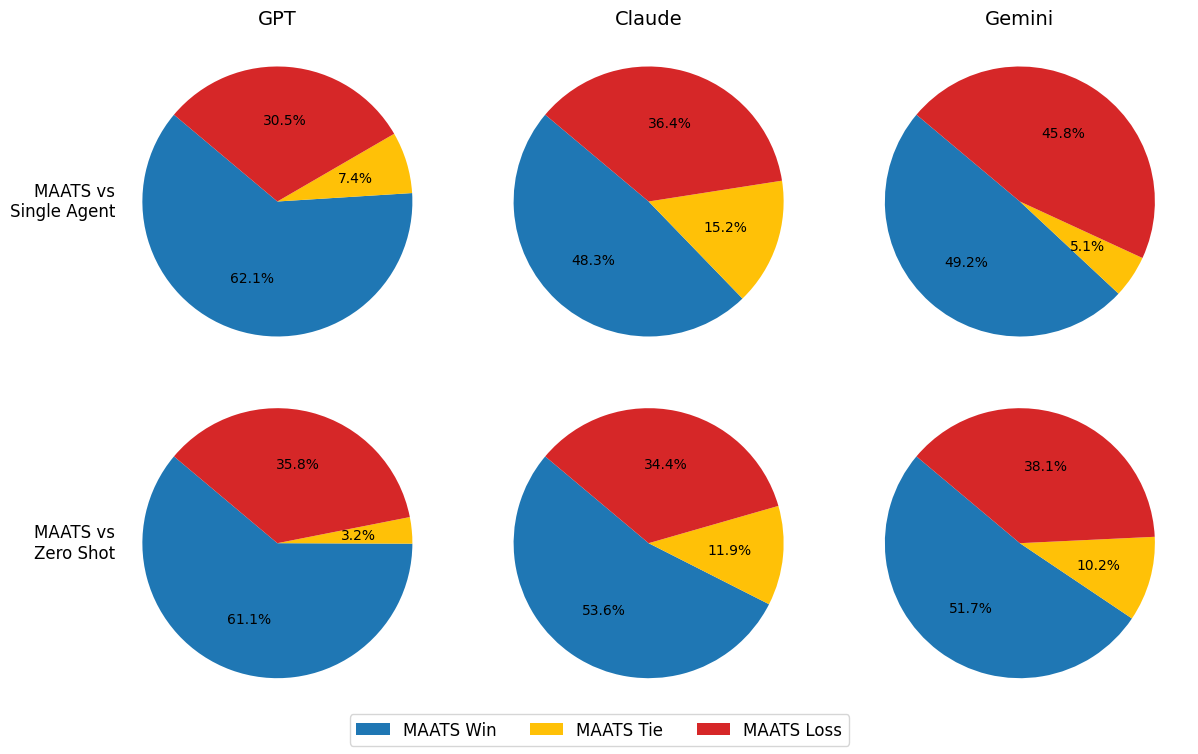}

    \caption{Human Evaluation Results: MAATS vs. Single-Agent and Zero-Shot in EN→ZH Translation }
    \label{fig:maats human ranking}
\end{figure}

\begin{figure*}[t!]

    \centering
    \includegraphics[width=\textwidth]{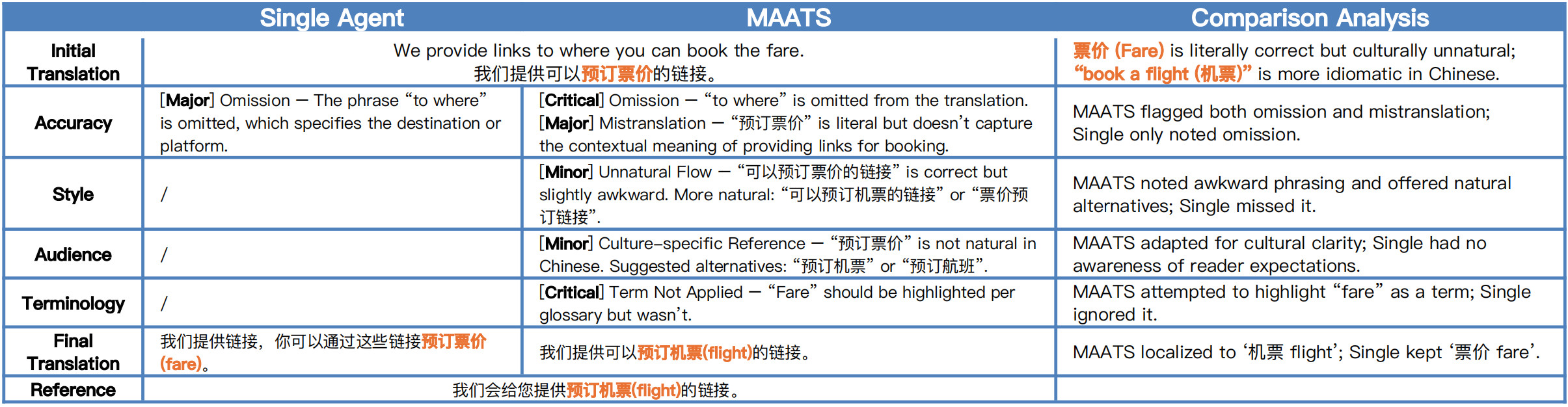}
    \begin{CJK*}{UTF8}{gbsn}
    \caption{MAATS Corrects Cultural and Contextual Errors in “Book the Fare. By flagging Accuracy, Style, Audience, and Terminology, “预定票价” was revised to the more natural and contextually appropriate “预定机票".}
    \label{fig:case_study_figure}
    \end{CJK*}
    
\end{figure*}

\noindent compared to 9,217 errors detected by the single-agent approach—an increase about 450\% (Fig.~\ref{fig:overall maats vs single}). In the ZH$\rightarrow$EN direction (Fig.~\ref{fig:zh_en maats vs single}), MAATS demonstrated greater sensitivity in identifying issues related to Accuracy, Style, Audience and Terminology, as shown in Fig. \ref{fig:case_study_figure}. Error counts across all evaluated language pairs are shown in Fig.~\ref{fig:appendix:error-by-language}. These results suggest that MAATS offers more opportunities for refinement.

\noindent\textbf{Performance Gains.} We compare three baselines: direct translation (zero-shot), single-agent refinement \cite{feng2024tearimprovingllmbasedmachine}, and our proposed MAATS approach. MAATS consistently achieves broader coverage and higher scores across metrics and directions. Of the LLMs compared, GPT-4o model showed the largest improvements, especially for German-to-English translations, with BLEU rising by +10.6 and COMET by + 8.7, according to Fig. ~\ref{fig:fig-gpt4o}. ANOVA test between approaches revealed that MAATS consistently outperformed baselines across most language pairs, with strongest gains on neural metrics like COMET and BLEURT (p < 0.001). Improvements were most pronounced in linguistically distant pairs such as EN$\leftrightarrow$JA, EN$\leftrightarrow$ZH and EN$\leftrightarrow$HE (see Fig.\ref{fig:anova-gemini}, \ref{fig:anova-claude},  \ref{fig:anova-gpt4o} \& Appendix \ref{statistical analysis}) and pairwise comparisons (Table \ref{tab:pairwise}).  \\




\begin{figure}[ht]
    \centering
    \includegraphics[width=0.48\textwidth]{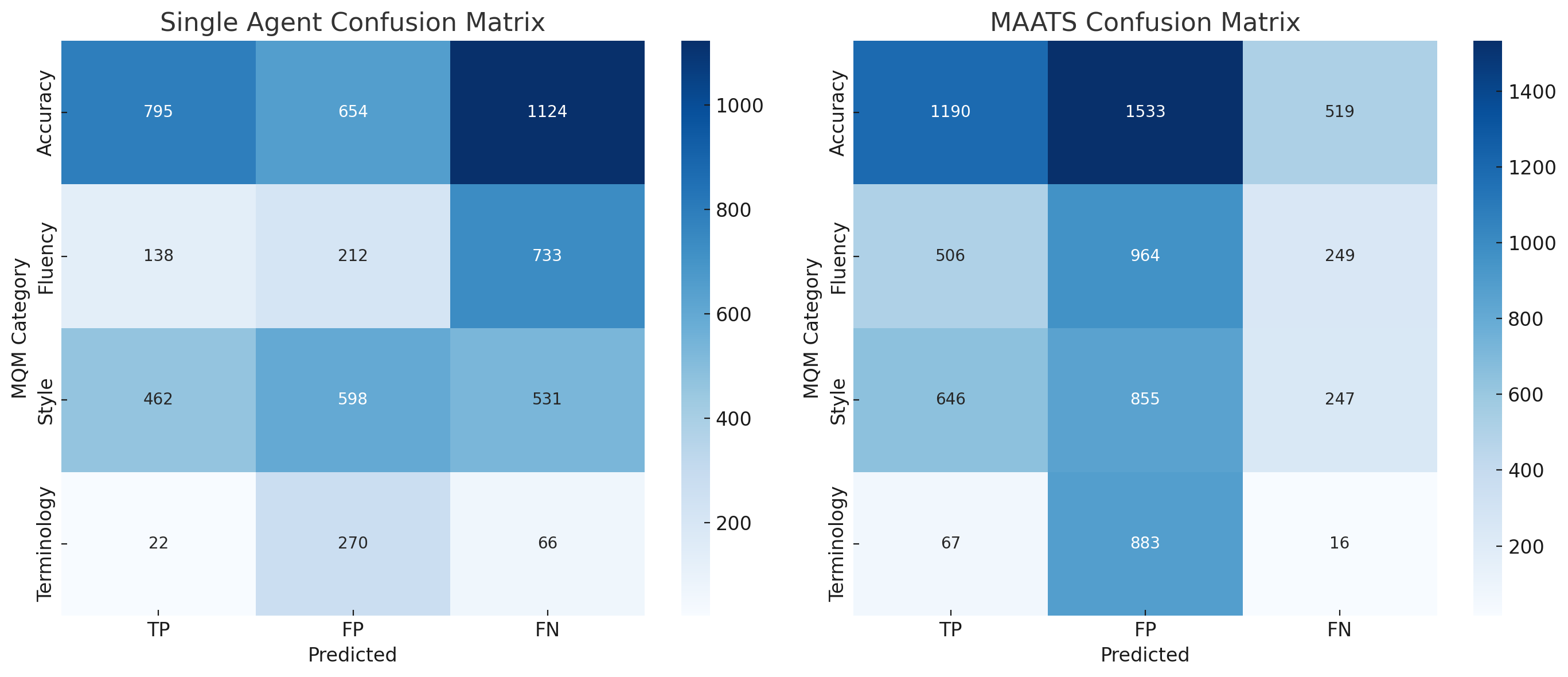}
    \caption{GPT-Based Confusion Matrix}
    \label{fig:confusion-gpt}
\end{figure}

\noindent\textbf{MAATS: Better Error Detection?} To validate MAATS’s error detection performance, we compared its outputs against human-labeled MQM reference data using confusion matrices. The GPT-based confusion matrix (Table \ref{fig:confusion-gpt}) shows that MAATS clearly outperforms the baseline across all major MQM categories. In Accuracy, it boosts true positives by 49.7\% and cuts false negatives by more than half. Fluency detection improves 3.7×, while Style sees a 40\% increase in TPs and a 53.5\% reduction in FNs. Although MAATS introduces more false positives, our analysis confirms that the majority of these are minor issues rather than major misjudgments (Comparison to other models see table \ref{tab:combined-confusion-matrix}). This aligns with prior studies suggesting that false positives in MT evaluations often reflect low-severity problems \cite{perrella2024correlationinterpretableevaluationmachine,freitag2021experts}. Overall, MAATS demonstrates greater sensitivity and alignment with human evaluation with more actionable translation issues.

\noindent\textbf{Do Humans Prefer MAATS?} To validate our automated evaluation, we conducted a human annotation experiment using English-to-Chinese translations from WMT 2023. A panel of three professional bilingual translators independently ranked outputs from MAATS, single-agent, and zero-shot approaches using MQM criteria. See appendix \ref{human ranking} for detailed experiment setup. As shown in Fig. \ref{fig:maats human ranking}, MAATS consistently outperformed both baselines. It achieved win rates of 62.1\%, 48.3\%, and 49.2\% against single-agent, and 61.0\%, 53.6\%, and 51.7\% against zero-shot across GPT, Claude, and Gemini, respectively. These human evaluation results confirm MAATS's effectiveness and reliability while aligning with our automated metrics.

\section{Discussion}

\begin{CJK*}{UTF8}{gbsn}
Beyond the primary evaluation, we conducted a qualitative meta-analysis of three case studies to compare different approaches. We revealed that MAATS consistently excels in three key areas: (1) multi-layered error analysis beyond surface issues, (2) detects omissions and subtleties across perspectives, and (3) enriches context to match target locale expectations. Shown in Fig. \ref{fig:case_study_figure}, MAATS flagged the unnatural literal translation “预定票价” and revised it to the idiomatic “预定机票.” It also uncovered cultural implications in “Car Guy Car” (see Table \ref{case study 1}), inferred implied actions in “I started a new shawl” (Table \ref{case study 3.1}),  and adjusted tone in “literature class hatred” (Table \ref{case study 3.2}) for classroom context. These examples demonstrate MAATS’s superior sensitivity to context and nuance (full comparisons in Appendix \ref{case study main }).
\end{CJK*}

In conclusion, MAATS bridges the gap between black-box LLMs applied to MT and human translation workflows by using interpretable MQM dimensions and modular roles. It shifts the focus from superficial fluency toward deeper semantic fidelity and contextual alignment, producing outputs that better reflect human standards. Its distributed architecture not only boosts translation performance but also introduces robustness and modularity—laying the groundwork for broader applications of multi-agent collaborative AI in structured NLP tasks.

\section{Limitations}

\begin{CJK*}{UTF8}{gbsn} 
While MAATS shows strong performance in error detection and correction, our meta-analysis reveals key limitations shared by both MAATS and Single-Agent systems, particularly in capturing emotional tone, pragmatic nuance, and rhetorical structure. These challenges stem from both model behavior and inherent constraints in the MQM framework. For instance, emotionally charged terms like “hatred” were mistranslated with excessive intensity “仇恨” (see Table \ref{case study 3.2}), missing the speaker’s intended tone. Sentences with emotional contrast, such as “Abe keeps his sorrows hidden and laughs easily,” were misinterpreted as lighthearted, failing to capture the ironic or concealed emotion. Moreover, rhetorical structures like repetition (e.g., “ads, ads, ads”) were either rigidly translated or replaced with fluent idioms, losing the original stylistic emphasis. These examples highlight that while MAATS excels in accuracy and fluency, it struggles with affective meaning conveyed through tone, irony, or form. Crucially, MQM lacks dimensions for evaluating emotional fidelity, speaker stance, or discourse-level intent, leaving such nuances under-annotated and uncorrected. Addressing these issues may require expanding MQM or integrating it with complementary methods like discourse modeling and affective computing to better reflect the full communicative intent in translation.
\end{CJK*}





\section{Ethical Implications}
While MAATS improves translation quality and interpretability, it introduces several ethical concerns. First, it inherits biases from underlying LLMs and may reinforce them through overcorrection, especially in culturally sensitive content. Second, MAATS's strong performance risks overreliance on automation, potentially sidelining human judgment in nuanced or affective contexts. Third, its replication of human workflows may contribute to labor displacement among translators and post-editors. Finally, its superior performance when translating into English reflects an imbalance rooted in pretraining data, potentially reinforcing linguistic hegemony. To mitigate these risks, MAATS should be deployed with human oversight, regular bias audits, and transparency in design. A deeper qualitative analysis will also further our understanding of mitigating biases in MT using MAATS. Future extensions could include fairness-aware prompts or an Ethics Evaluator Agent to flag sensitive content.


\clearpage
\appendix
\section{Baseline and MAATS Implementation Details}
\label{appendix:A}

\paragraph{Single-Agent Baseline}
We construct a strong baseline where each model uses a self-refinement strategy \cite{feng2024tearimprovingllmbasedmachine}. Concretely, after the model produces its initial translation, we prompt it with a message like: \textit{“Please review your translation above. Identify any errors or improvements and then provide a corrected translation.”} This makes the model critique and refine its own output in one go. We found that one round of self-refinement is usually sufficient; more rounds yield diminishing returns and sometimes oscillations (e.g., the model changes a phrase one way, then reverts it). Therefore, the Single-Agent system output we use is the result after one self-refinement pass. This baseline is considerably stronger than taking the first draft translation, as it gives the model a chance to fix obvious mistakes. It is essentially an ablated version of MAATS, where the same model plays all roles internally (but without the explicit structure of MQM categories—the model is free to decide what to fix). In rare cases where the model refused or made no changes on the refinement prompt, we take the initial output as final.

\paragraph{MAATS Implementation Details}
For MAATS, we crafted prompts for each agent role. The translator prompt is straightforward (\textit{“Translate the following.”}). The evaluator prompt is crucial: for each MQM category, the agent identifies all errors with justification and severity level. We included examples in the prompt to guide annotation format (e.g., showing a sample error report). The editor agent then consolidates all annotations from the evaluator agents to produce a refined translation. We ensured that the evaluator outputs were included in the editor’s context.

One challenge was managing context length, as including the source, translation, and full list of errors can become long. However, with GPT-4 and other models supporting extended context windows, this was manageable for our input size (each input was typically under 1000 tokens). To ensure consistency and reduce randomness during refinement, we kept the model temperature low (between 0 and 0.3), aiming for deterministic improvements rather than alternate phrasings.

\section{Translation Task Design}
\label{appendix:translation-task}

To evaluate the MAATS approach, we designed experiments across multiple language directions and translation paradigms. The evaluation covered bidirectional translation between English and six other languages: German, Hebrew, Japanese, Russian, Chinese, and Arabic—resulting in a total of eleven translation directions.

For each direction, we used a test set of approximately 200 sentences, drawn from the WMT 2023 and WMT 2024 datasets. These source texts span a range of content, including factual, idiomatic, and technical sentences from news and general domain parallel corpora. The dataset size was chosen to balance feasibility with sufficient coverage for robust metric evaluations and fine-grained error analysis.

\section{Human Ranking Evaluation Setup}
\label{human ranking}

To complement our automated evaluation, we conducted a human annotation experiment for the English-to-Chinese translation task using samples from WMT 2023. We recruited three professional bilingual translators with diverse industry credentials: \\
\textbf{Annotator A}: Holds a master’s degree in translation and has over 10 years of experience. Certified at the Chinese national Level II standard in Translation.\\
\textbf{Annotator B}: Holds a Level III Chinese national certification in Translation with 5 years of professional experience.\\
\textbf{Annotator C}: Holds a graduate degree in translation, certified at Level II, with 3 years of industry experience.

\begin{figure}[ht]
    \centering
    \includegraphics[width=0.48\textwidth]{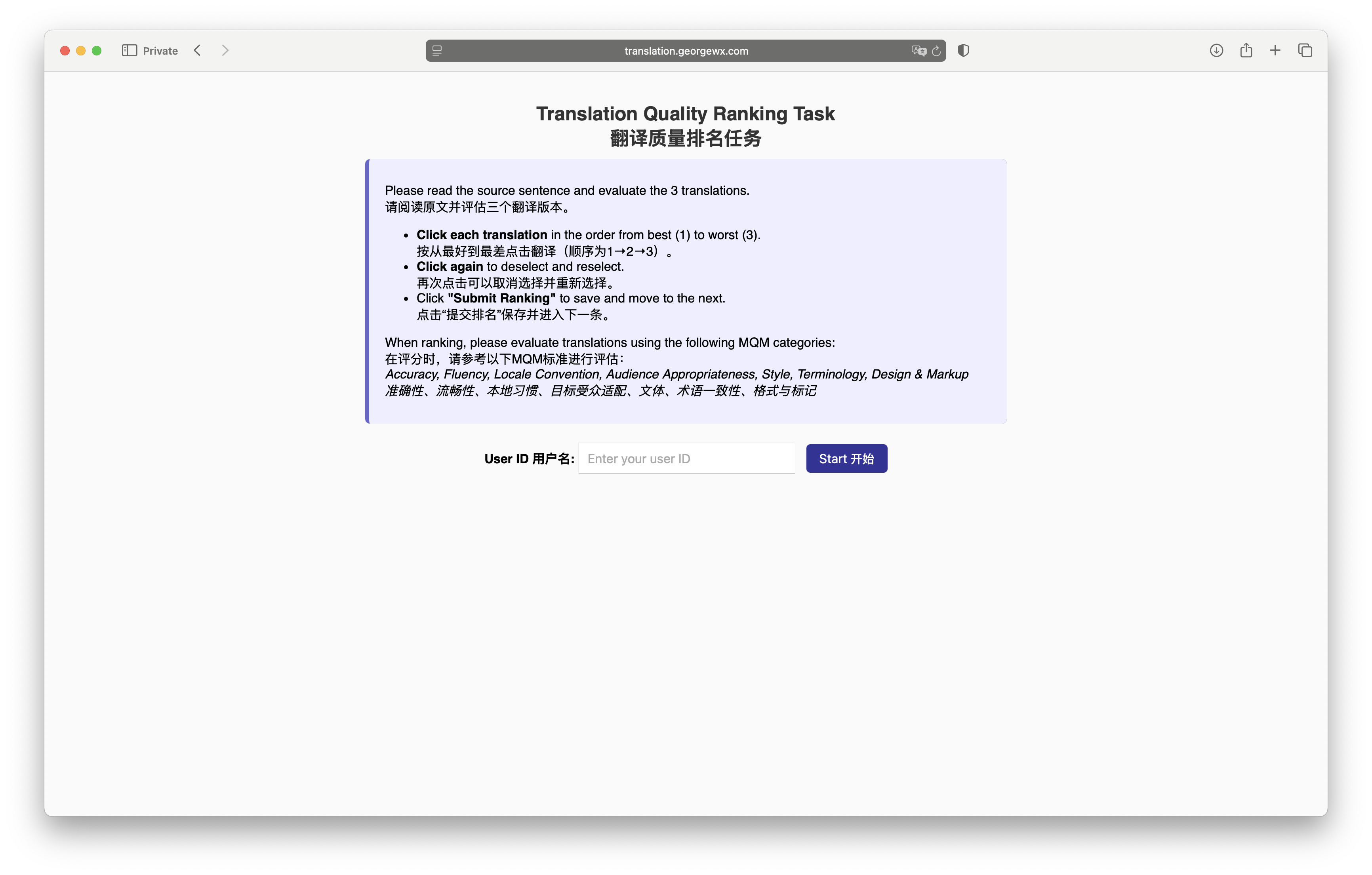}
    \caption{Translation Quality Ranking Interface used in the human evaluation experiment.}
    \label{fig:main-interface}
\end{figure}

All annotators confirmed familiarity with MQM (Multidimensional Quality Metrics) and were instructed to evaluate translations based on key MQM dimensions, as shown in Fig.~\ref{fig:main-interface}.

\begin{figure}[ht]
    \centering
    \includegraphics[width=0.48\textwidth]{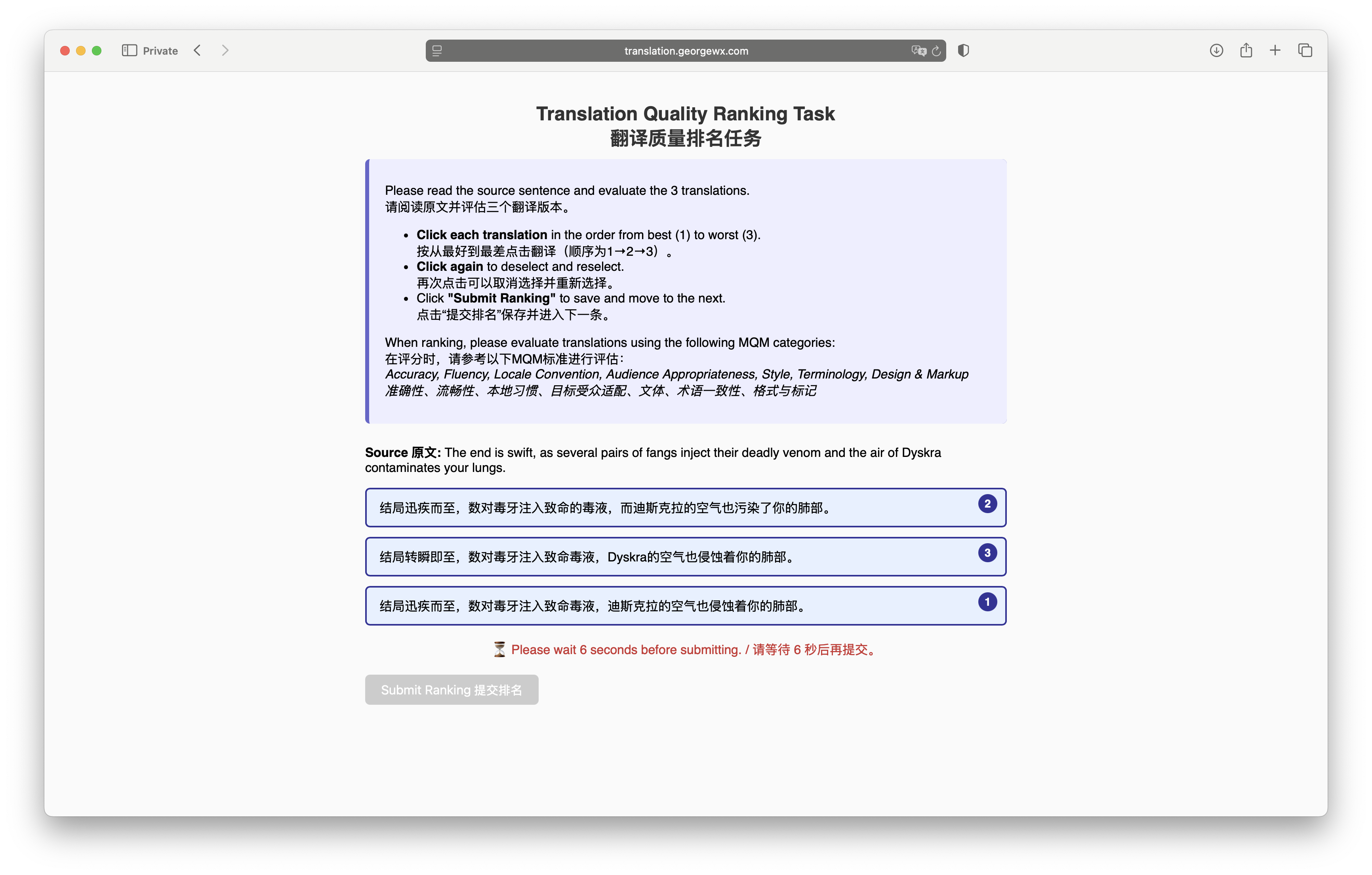}
    \caption{Web-based ranking interface for human evaluation.}
    \label{fig:ranking-interface}
\end{figure}

We developed a web-based interface for the ranking task (see Fig.~\ref{fig:ranking-interface}), where annotators reviewed each source sentence and ranked the three anonymized system outputs (MAATS, single-agent, and zero-shot) from best (1) to worst (3). Each annotator evaluated approximately 150 translation samples per LLM system, resulting in a total of 450 ranked examples across the three models.

Final rankings were aggregated using the \textit{Borda count} method, a positional voting system in which rankings are assigned scores (1st = 2 points, 2nd = 1 point, 3rd = 0 points). The system with the highest total score across annotators was considered the preferred output~\cite{emerson2013original}. This approach enabled consistent tie-breaking and fair consensus-based evaluation.

\section{Statistical Analysis}
\label{statistical analysis}
\begin{figure}[H]
    \centering
    \includegraphics[width=\linewidth]{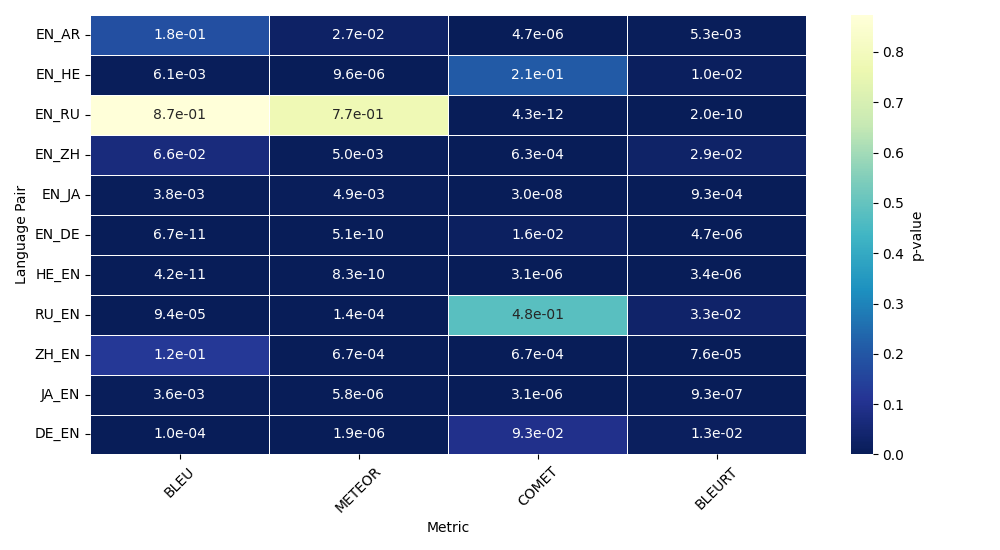}
    \caption{ANOVA Significance Heatmap with Claude Under MAATS}
    \label{fig:anova-gemini}
\end{figure}

\begin{figure}[H]
    \centering
    \includegraphics[width=\linewidth]{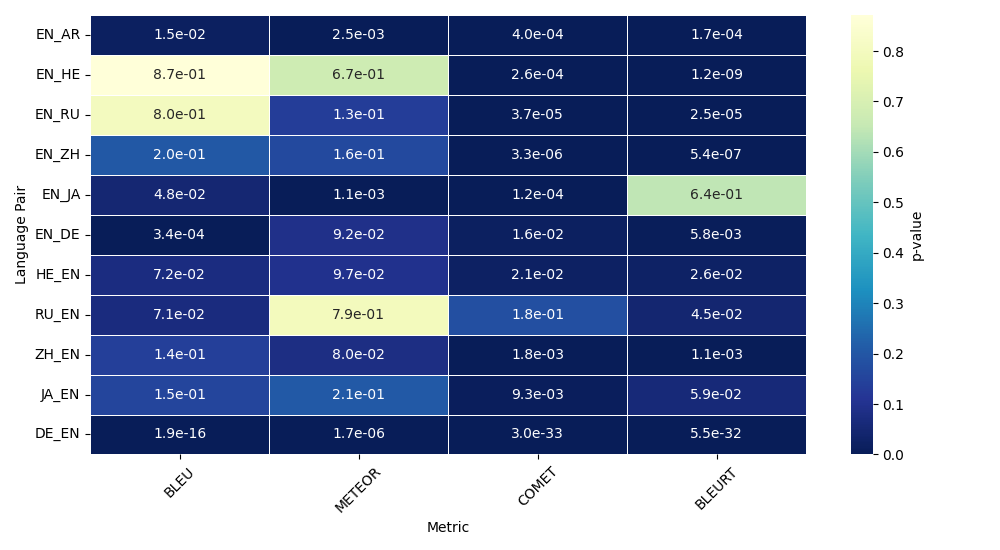}
    \caption{ANOVA Significance Heatmap with GPT Under MAATS}
    \label{fig:anova-claude}
\end{figure}

\begin{figure}[H]
    \centering
    \includegraphics[width=\linewidth]{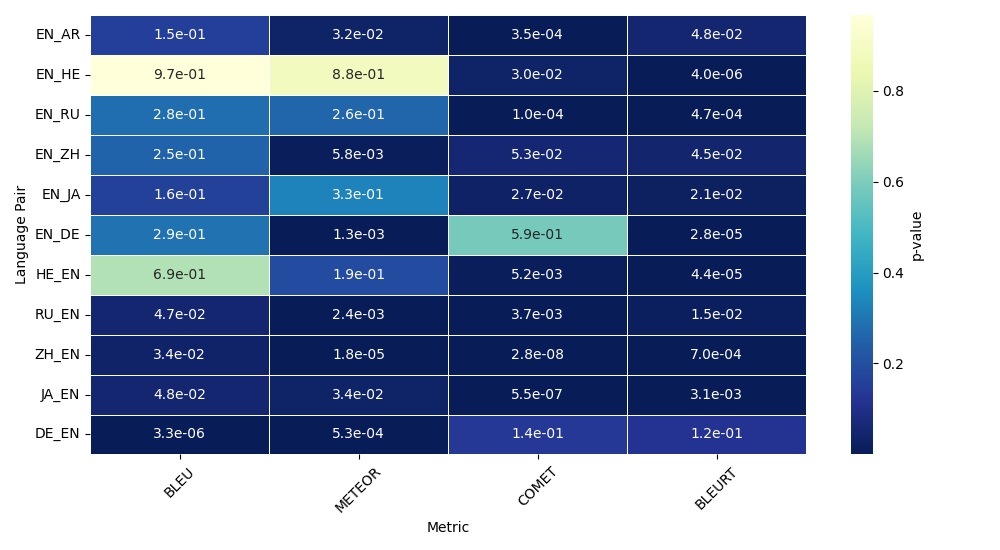}
    \caption{ANOVA Significance Heatmap with Gemini Under MAATS}
    \label{fig:anova-gpt4o}
\end{figure}

Fig.\ref{fig:anova-gemini}, Fig.\ref{fig:anova-claude} and Fig.\ref{fig:anova-gpt4o}   present ANOVA results across evaluation metrics, language pairs, and LLMs. COMET, a neural metric aligned with human judgment, shows the strongest and most consistent significance across models and languages. This confirms MAATS’s ability to enhance semantic accuracy and contextual coherence. BLEURT also yields robust significance. In contrast, METEOR achieves moderate significance in roughly two-thirds of cases, while BLEU—heavily reliant on surface n-gram overlap—shows significance in fewer than half. This implies that MAATS excels in deeper semantic and stylistic refinement rather than superficial lexical improvements. Model-wise, Claude-3-haiku benefits the most from MAATS, with consistent significance across all metrics and pairs. Gemini-2-flash exhibits directional asymmetry: improvements are limited when translating from English but substantial when translating into English. For GPT-4o, lexical metric gains are limited—likely due to its already fluent outputs. Neural metrics still show significant improvements pointing MAATS’s role in enhancing subtle semantic fidelity.

\section{Directional Asymmetry}
\label{directional}

\begin{figure}[t]
    \centering
    \includegraphics[width=\linewidth]{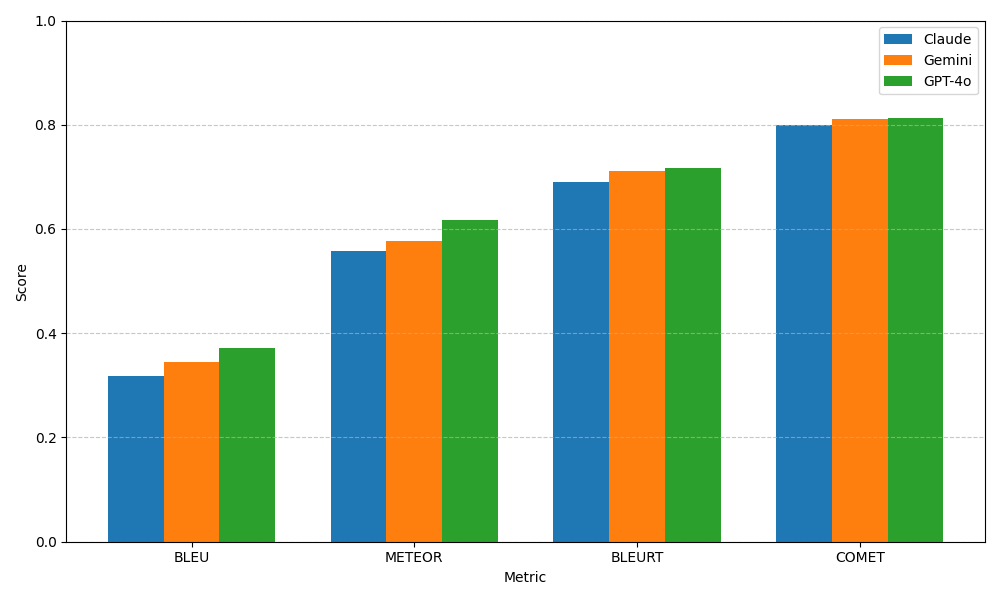}
    \caption{MAATS Performance Across Models on German to English}
    \label{fig:de en}
\end{figure}

\begin{figure}[t]
    \centering
    \includegraphics[width=\linewidth]{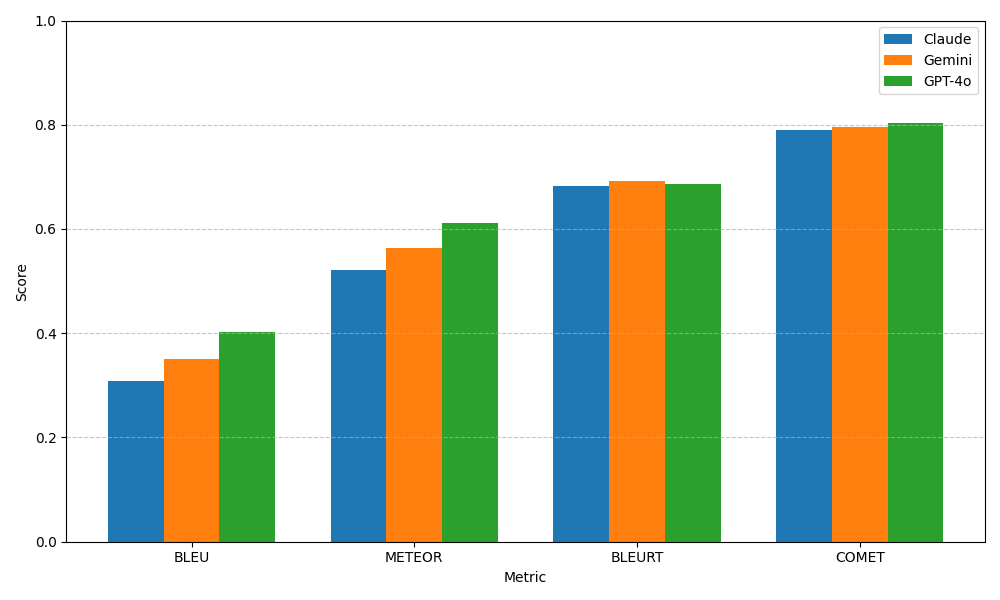}
    \caption{MAATS Performance Across Models on English to German}
    \label{fig:en de}
\end{figure}

The directional asymmetry in MAATS performance is clearly illustrated in Fig. \ref{fig:de en}  and \ref{fig:en de}, which present results for German to English and English to German translation tasks, respectively. Across all metrics, notably BLEU and METEOR, translations into English consistently outperform those into German. This trend highlights a systemic advantage when English serves as the target language.

Two key factors likely contribute to this disparity. First, LLMs are typically pretrained with greater English-language representation, enhancing baseline translation quality and providing more consistent inputs for MAATS agents. Second, MQM categories such as Style and Audience Appropriateness are more effectively addressed in English by current LLM evaluators and refiners. This asymmetry extends beyond the German-English pair. Similar patterns are observed in other language pairs (e.g., Chinese, Japanese, Russian, Hebrew, Arabic), where MAATS yields more substantial improvements when translating into English. This suggests a generalizable trend, driven by both pretrained model bias and agent optimization.

\section{Case Studies}
MAATS outperforms Single Agent not only in identifying culturally nuanced content, but more importantly in performing multi-layered error analysis, detecting omissions across perspectives, and contextually enriching translations to align with target locale expectations. This conclusion is supported by three in-depth case studies, each illustrating a distinct but interrelated strength of the Multi-Agent system (MAATS) in contrast to the Single Agent.

\label{case study main }
\subsection{Case Study 1: “Car Guy Car” – A Cultural Reference Missed in Classification}
\label{case study 1 analysis}

The first case focuses on MAATS’s ability to analyze errors more deeply and classify them more precisely. In Table \ref{case study 1}, both systems initially failed to correctly translate the culturally specific phrase “Car Guy Car”, a term that requires domain and contextual understanding. However, their error annotations diverged significantly. The Single Agent categorized the issue under Style (Major) and described it merely as an “Awkward Translation”, offering a vague and surface-level explanation with no actionable insights. In contrast, MAATS provided a much more structured and informative analysis, flagging the same issue under Locale (Minor), Audience Appropriateness (Major), and Terminology (Critical). These annotations collectively pointed out that the expression was not only awkward but also culturally inappropriate and terminologically significant. 
Although both systems ultimately corrected the error in their final outputs, the Single Agent’s annotation reflected a limited understanding of the root problem, which we define as a Primary Error—a deep, structural misalignment with the cultural context—versus a Secondary Error, which is the surface awkwardness it creates. The Single Agent tended to catch only secondary manifestations, while MAATS consistently identified primary error causes, offering a better foundation for meaningful revision.

\subsection{Case Study 2: “Book the Fare” – MAATS Detects Omissions Across Perspectives}
\label{case study 2 analysis}

\begin{CJK*}{UTF8}{gbsn} 

The second case, Table \ref{case study 2}, illustrates MAATS’s ability to detect errors that the Single Agent completely overlooks, and to do so across multiple dimensions. The source phrase “Book the fare” was literally translated by both systems as “预定票价”, which is unnatural in Chinese—where one would typically say “预定机票” (book a flight). The Single Agent provided no annotation at all, thereby missing the issue entirely. In contrast, MAATS flagged the phrase across four categories:
\end{CJK*} \\ \textbf{Accuracy (Major)} – stating that the translation fails to capture the contextual meaning;\\
\textbf{Audience Appropriateness (Minor)} – noting the expression as culturally inappropriate;\\
\textbf{Style (Minor)} – describing it as correct but awkward;\\
\textbf{Terminology (Critical)} – highlighting “fare” as a technical term needing special attention.

\begin{CJK*}{UTF8}{gbsn}
It’s worth noting that the terminology annotation itself was not fully accurate, as it emphasized highlighting “fare” as a term but did not correct its misinterpretation as “票价.” This introduced a degree of internal conflict among annotations, yet MAATS still arrived at the correct fix—translating the phrase as “预定机票.” This case shows that MAATS not only detects more errors but also demonstrates robust decision-making in the face of conflicting signals. The system’s redundancy—flagging the issue multiple times from different perspectives—may serve as a self-correcting mechanism, compensating for occasional misclassifications.

\end{CJK*}
\subsection{Case Study 3}

\begin{CJK*}{UTF8}{gbsn}
The third case, Table \ref{case study 3.1} and Table \ref{case study 3.2}, highlights MAATS’s ability to enrich translations contextually based on cultural and linguistic expectations of the target locale. The phrase “I started a new shawl” was literally translated with “开始” (to start), producing a grammatically acceptable but culturally unnatural expression in Chinese. A more idiomatic translation would be “我开始织” (I started knitting) or “我开始制作” (I started making), which conveys the intended meaning more clearly in the target language. 

The Single Agent marked the sentence as No Error, failing to recognize the subtle mismatch. In contrast, MAATS flagged the sentence under Style (Minor) with the explanation “It is correct but not natural.” This demonstrates MAATS’s sensitivity to contextual and cultural norms, and its ability to infer omitted but expected content from surrounding context. Rather than simply correcting errors, MAATS engages in contextual enrichment—adjusting literal translations to better align with native phrasing and reader expectations.

A second example that supports this contextual sensitivity is found in Sentence 448, which includes the phrase “literature class hatred.” The term “hatred” was initially translated as “仇恨” by both Single and Multi-Agent systems. However, only MAATS revised it to “厌恶”, a word that better fits the intended nuance in context.

\end{CJK*}
Although “hatred” typically implies an extreme emotion, in this specific context—a classroom reflection on course preferences—the term should express strong dislike rather than literal hatred. The latter would be too emotionally intense and misleading in Chinese, especially considering the contrastive structure in the latter half of the sentence, where the speaker says, “but I enjoyed reading.” This contrast indicates that the speaker is not expressing ideological or moral condemnation, but simply stating a relative lack of enjoyment.

MAATS captured this nuance by marking the phrase in three categories, all with Major severity: \\
\textbf{Accuracy} – recognizing the semantic mismatch; \\
\textbf{Audience Appropriateness} – identifying the cultural inappropriateness of using such an emotionally charged word; \\ 
\textbf{Style} – noting that the tone felt jarring or unnatural. \\
In contrast, the Single Agent not only failed to revise the word, but in its Accuracy annotation, it incorrectly suggested that “hatred” should be made stronger, proposing “severe hatred” as a more accurate translation. This recommendation demonstrates a complete misreading of the sentence’s tone and communicative intent, further emphasizing the Single Agent’s lack of contextual grounding.


\newpage
\twocolumn

\begin{figure*}[t]
    \centering

    \begin{minipage}[t]{0.48\textwidth}
        \centering
        \includegraphics[width=\linewidth]{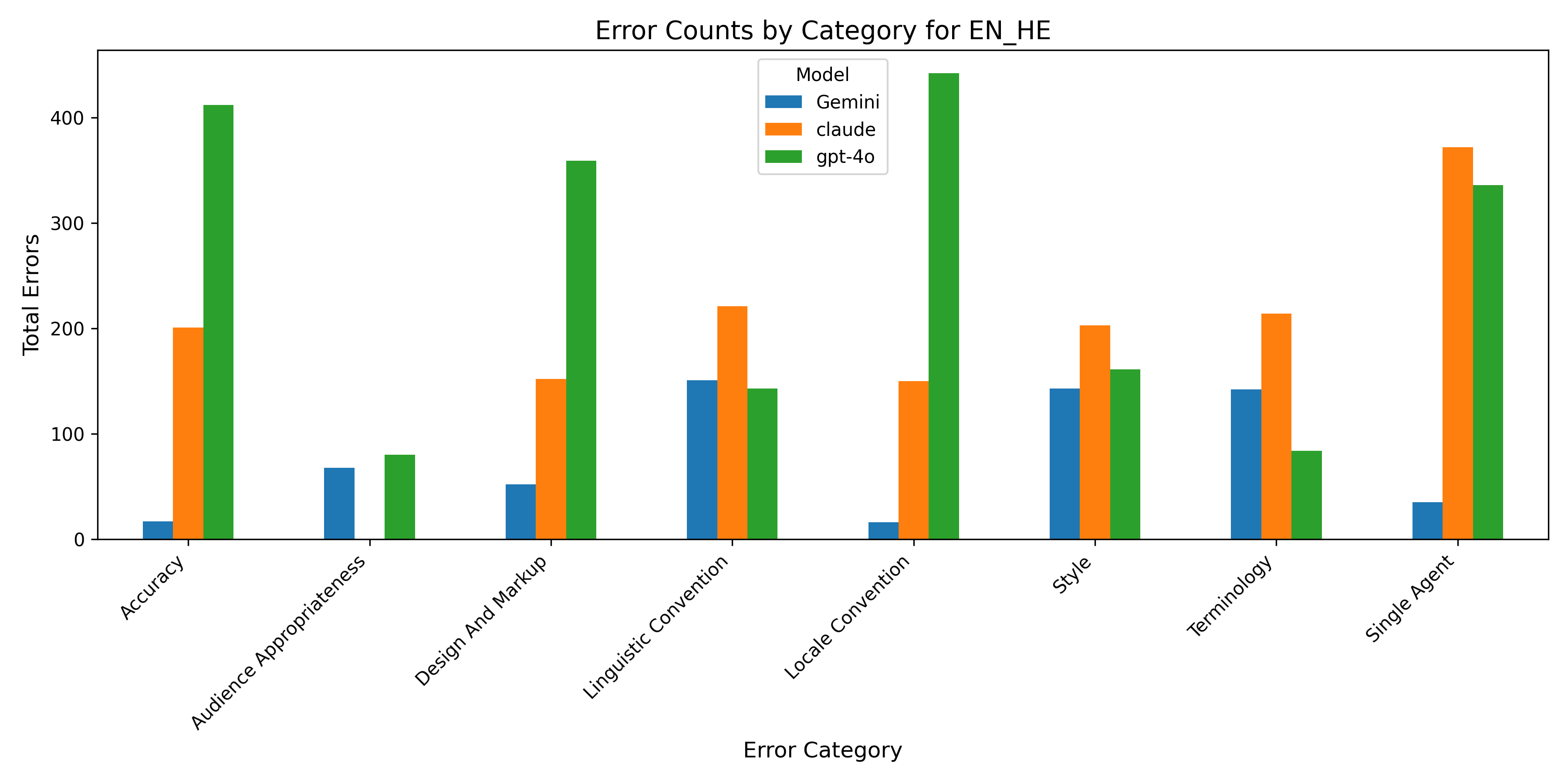}
        \caption*{EN$\rightarrow$HE}
    \end{minipage}
    \hfill
    \begin{minipage}[t]{0.48\textwidth}
        \centering
        \includegraphics[width=\linewidth]{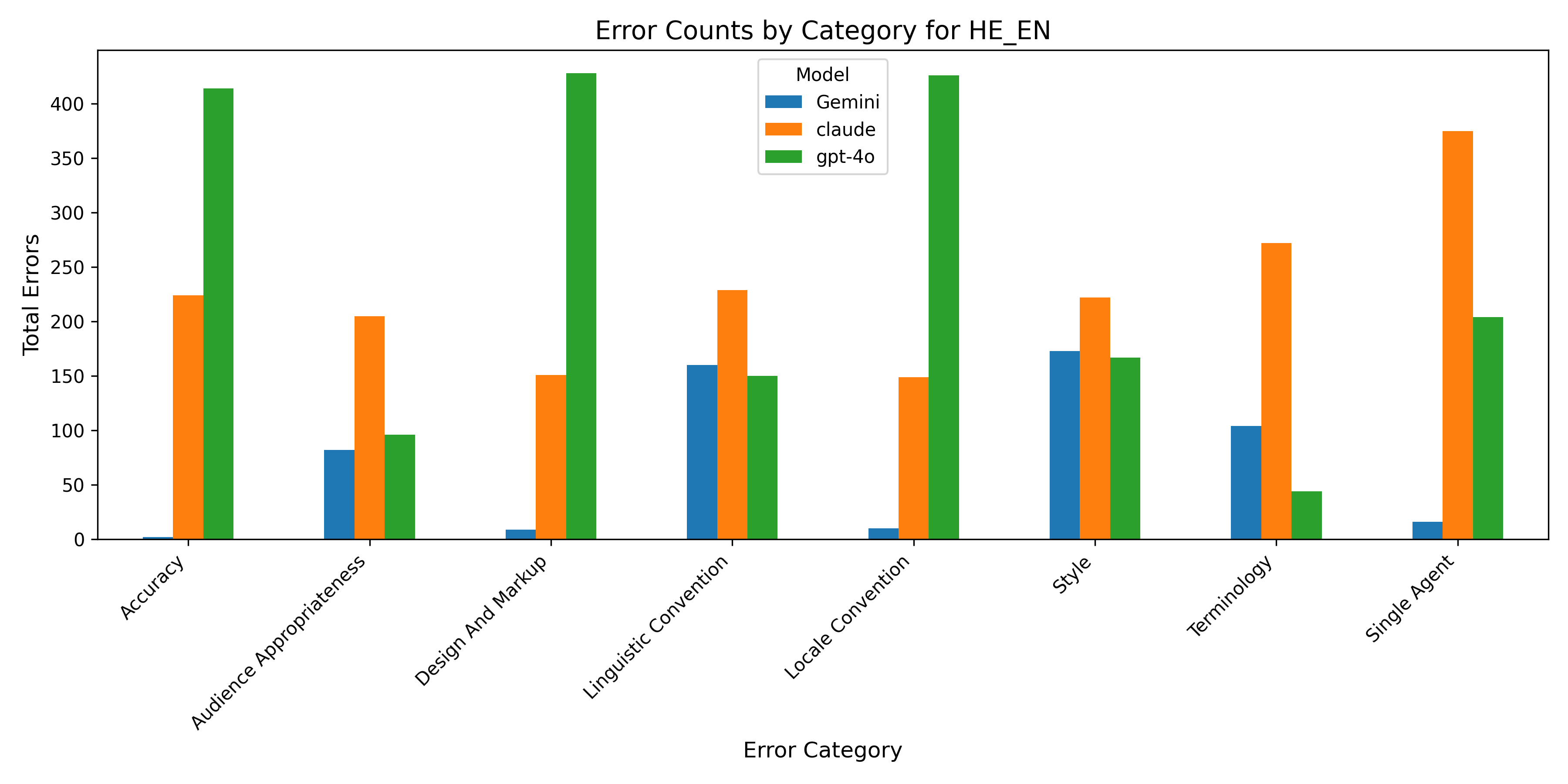}
        \caption*{HE$\rightarrow$EN}
    \end{minipage}

    \vspace{0.25em}

    \begin{minipage}[t]{0.48\textwidth}
        \centering
        \includegraphics[width=\linewidth]{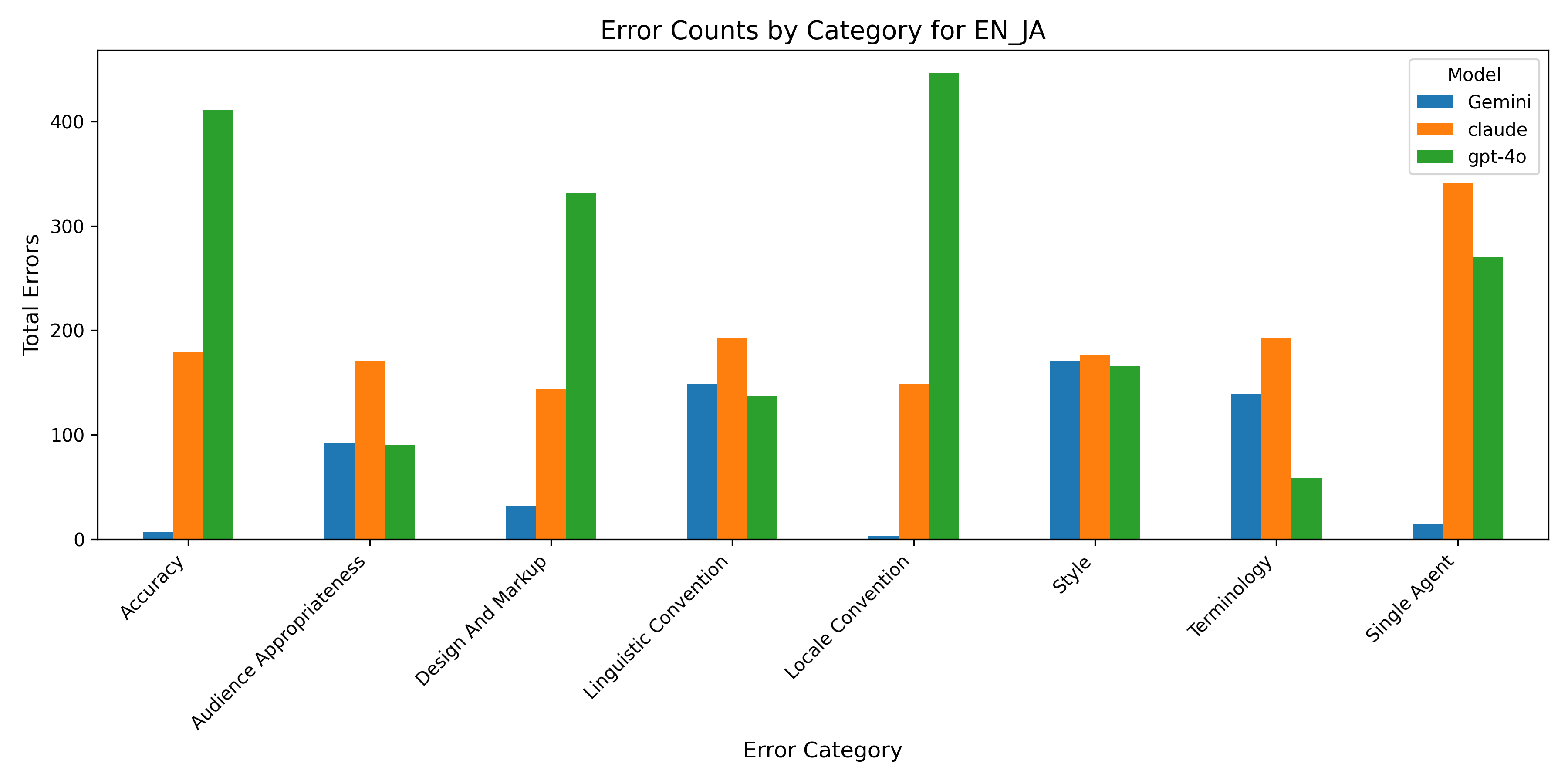}
        \caption*{EN$\rightarrow$JA}
    \end{minipage}
    \hfill
    \begin{minipage}[t]{0.48\textwidth}
        \centering
        \includegraphics[width=\linewidth]{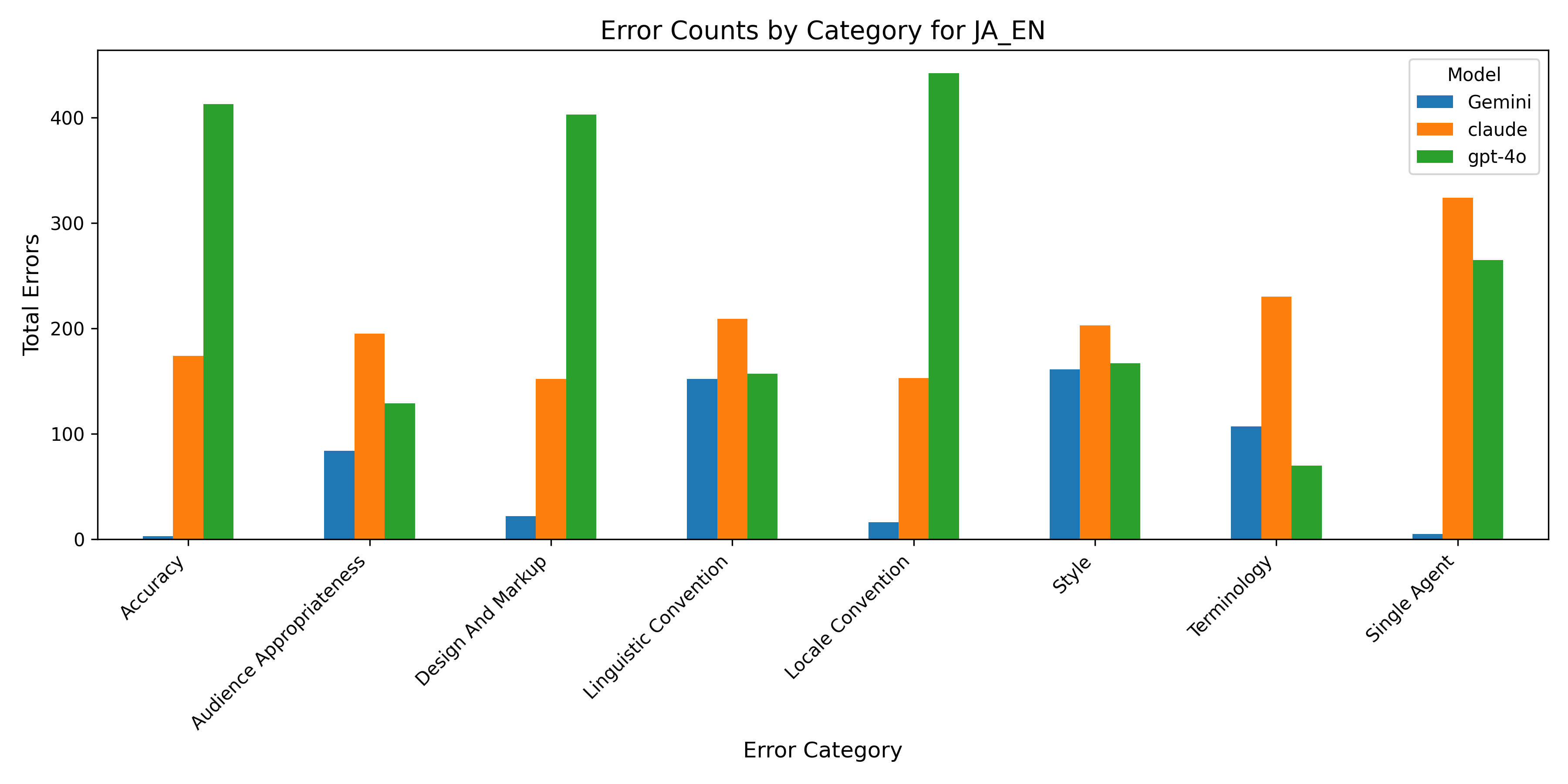}
        \caption*{JA$\rightarrow$EN}
    \end{minipage}

    \vspace{0.25em}

    \begin{minipage}[t]{0.48\textwidth}
        \centering
        \includegraphics[width=\linewidth]{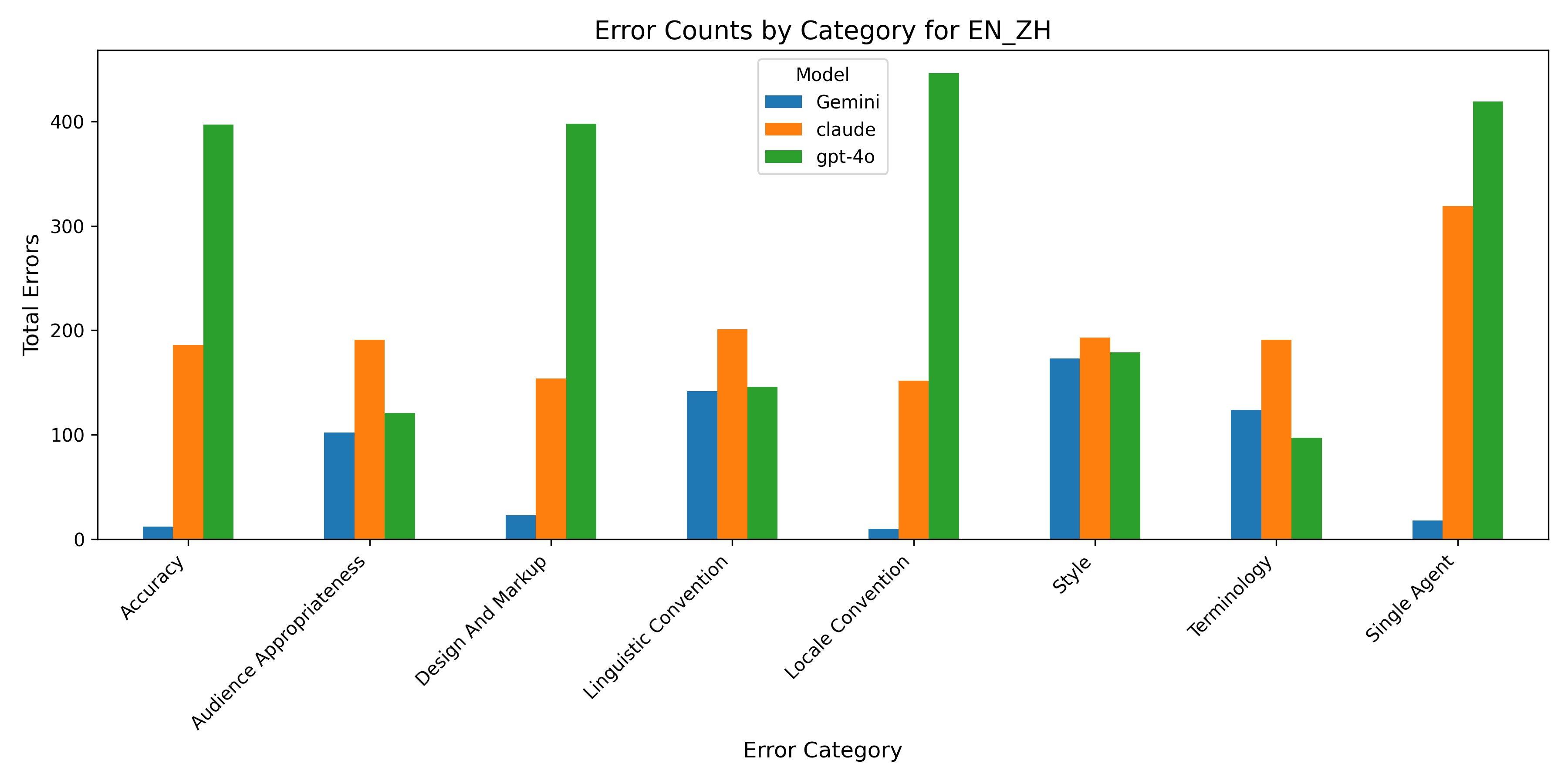}
        \caption*{EN$\rightarrow$ZH}
    \end{minipage}
    \hfill
    \begin{minipage}[t]{0.48\textwidth}
        \centering
        \includegraphics[width=\linewidth]{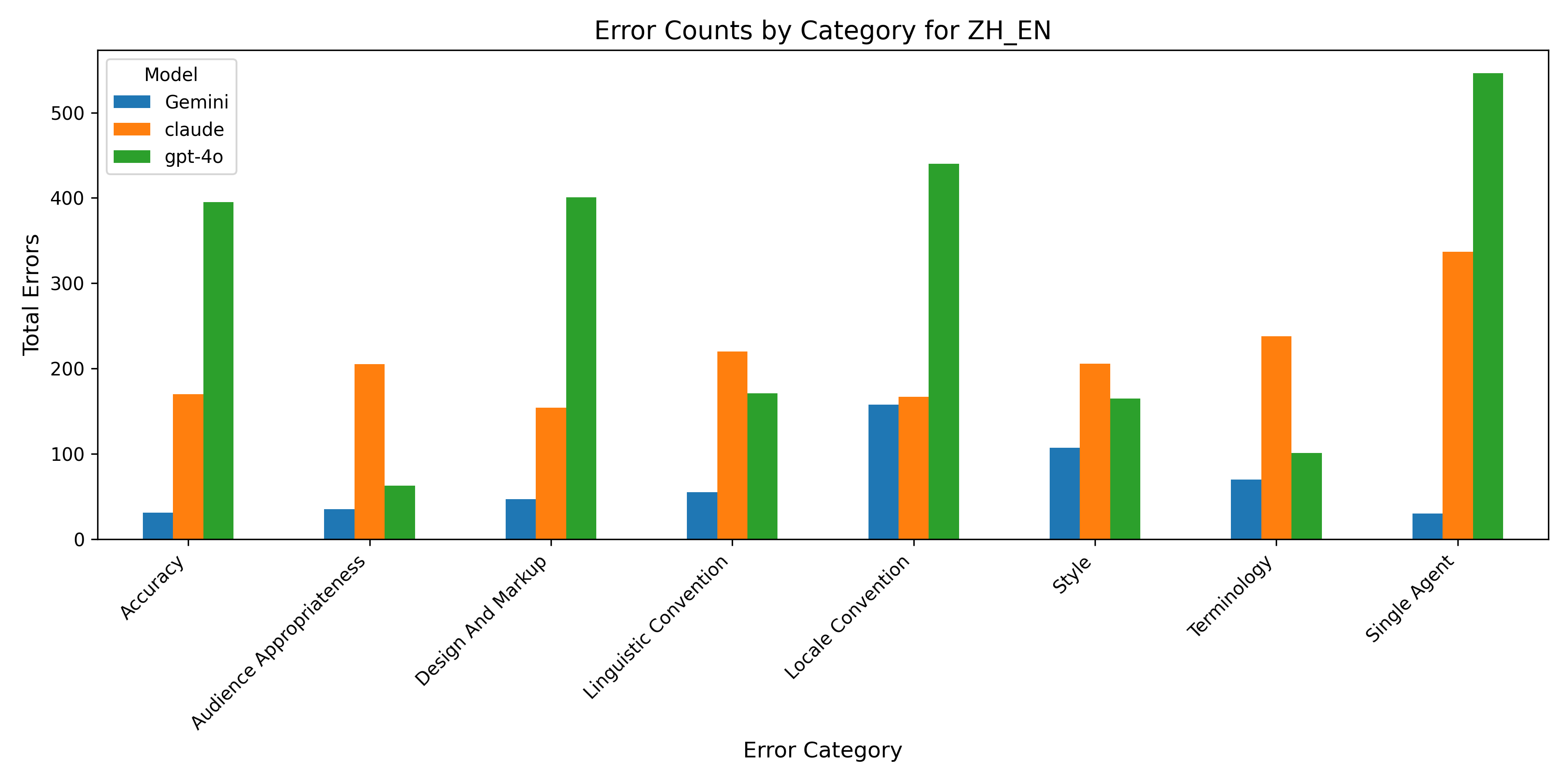}
        \caption*{ZH$\rightarrow$EN}
    \end{minipage}

    \vspace{0.25em}

    \begin{minipage}[t]{0.48\textwidth}
        \centering
        \includegraphics[width=\linewidth]{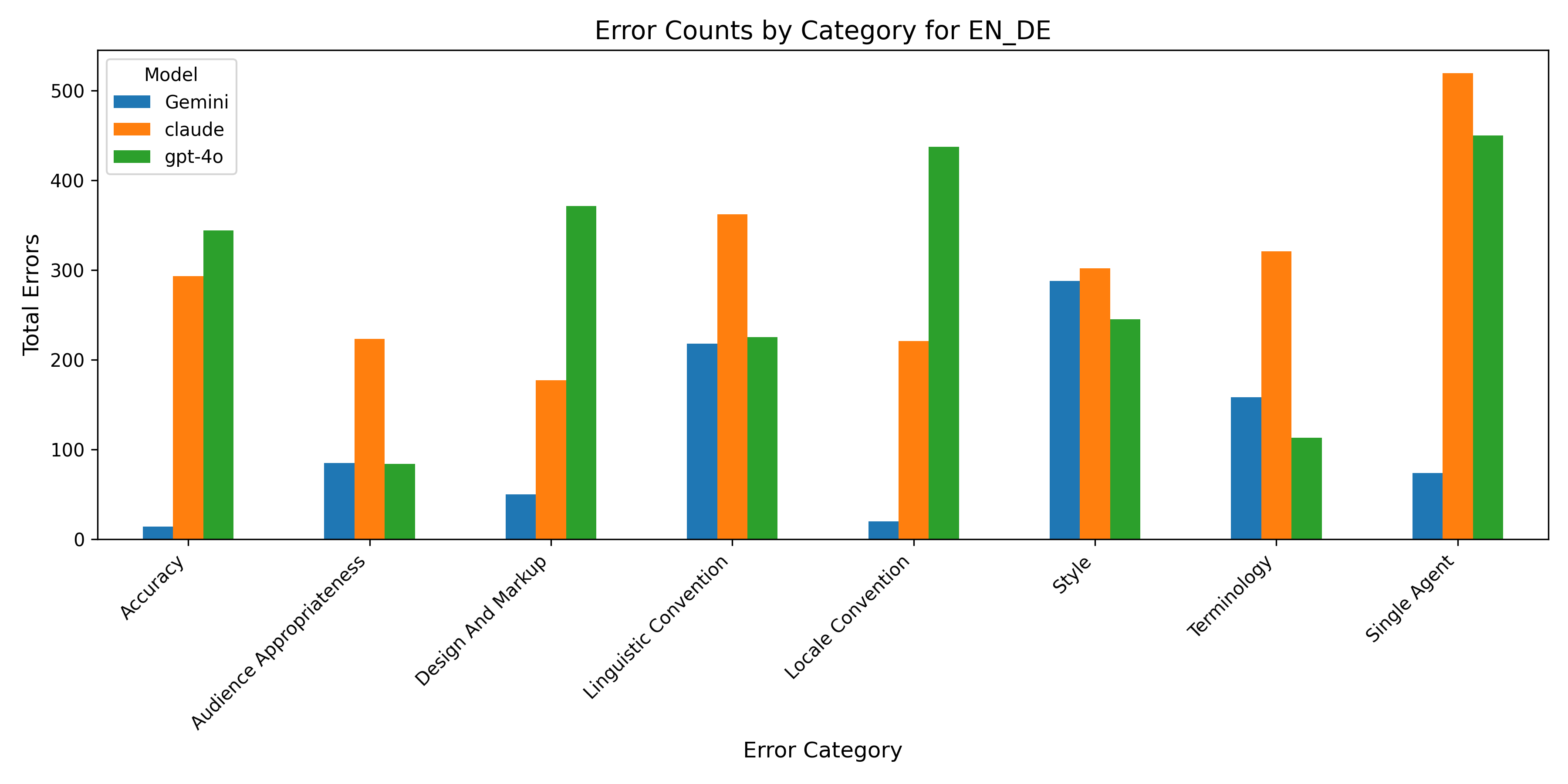}
        \caption*{EN$\rightarrow$DE}
    \end{minipage}
    \hfill
    \begin{minipage}[t]{0.48\textwidth}
        \centering
        \includegraphics[width=\linewidth]{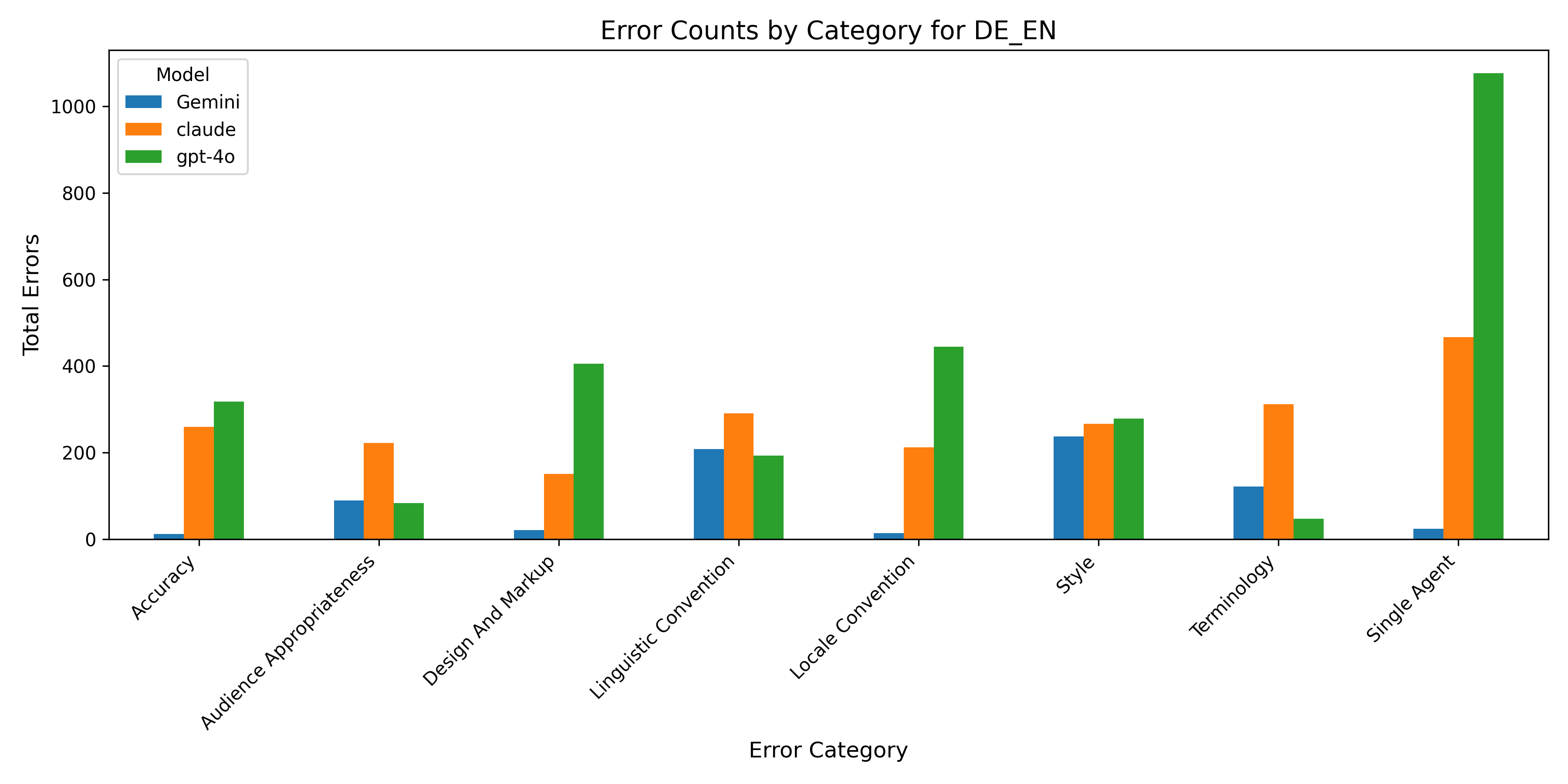}
        \caption*{DE$\rightarrow$EN}
    \end{minipage}

    \vspace{0.25em}

    \begin{minipage}[t]{0.48\textwidth}
        \centering
        \includegraphics[width=\linewidth]{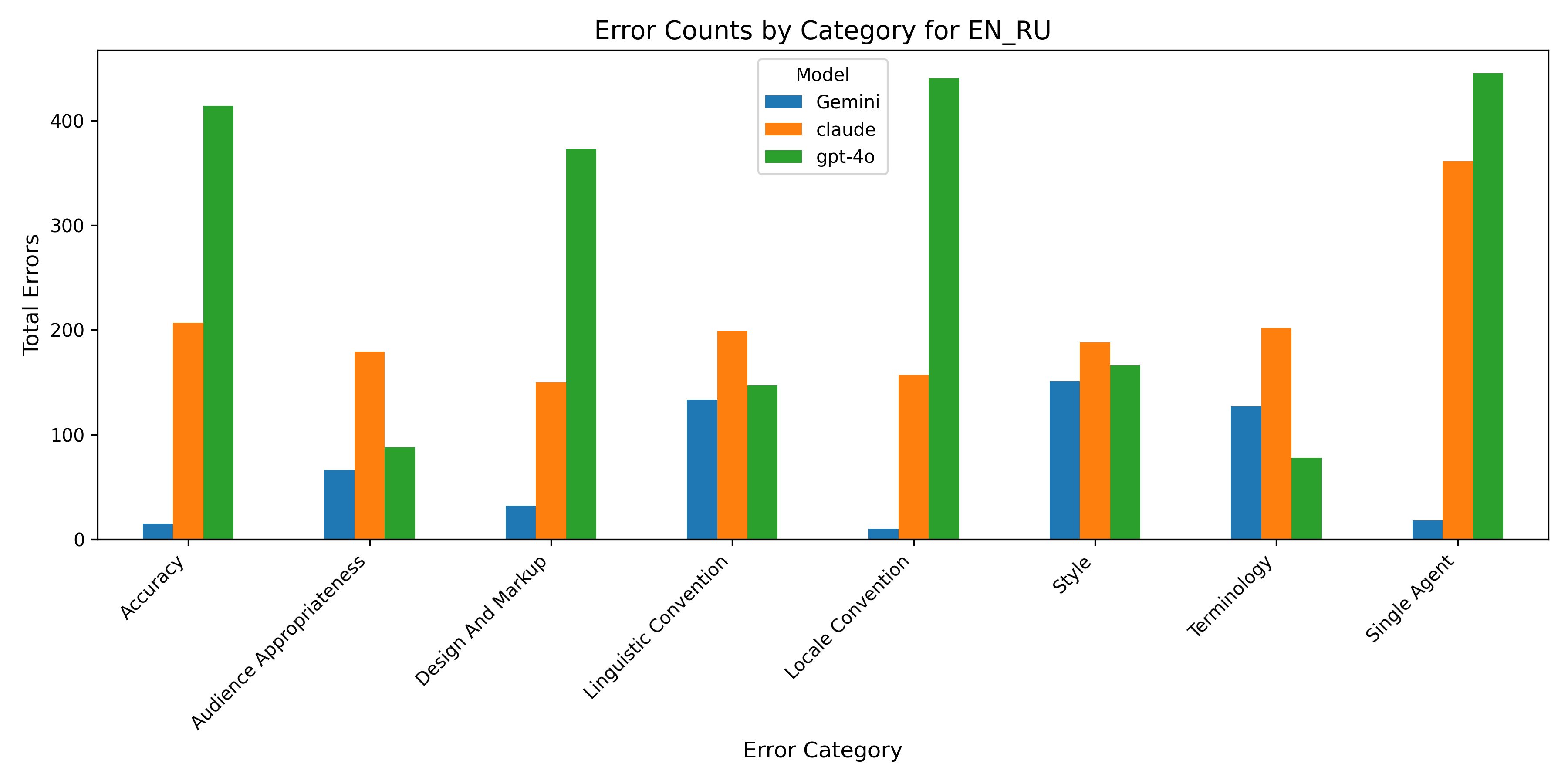}
        \caption*{EN$\rightarrow$RU}
    \end{minipage}
    \hfill
    \begin{minipage}[t]{0.48\textwidth}
        \centering
        \includegraphics[width=\linewidth]{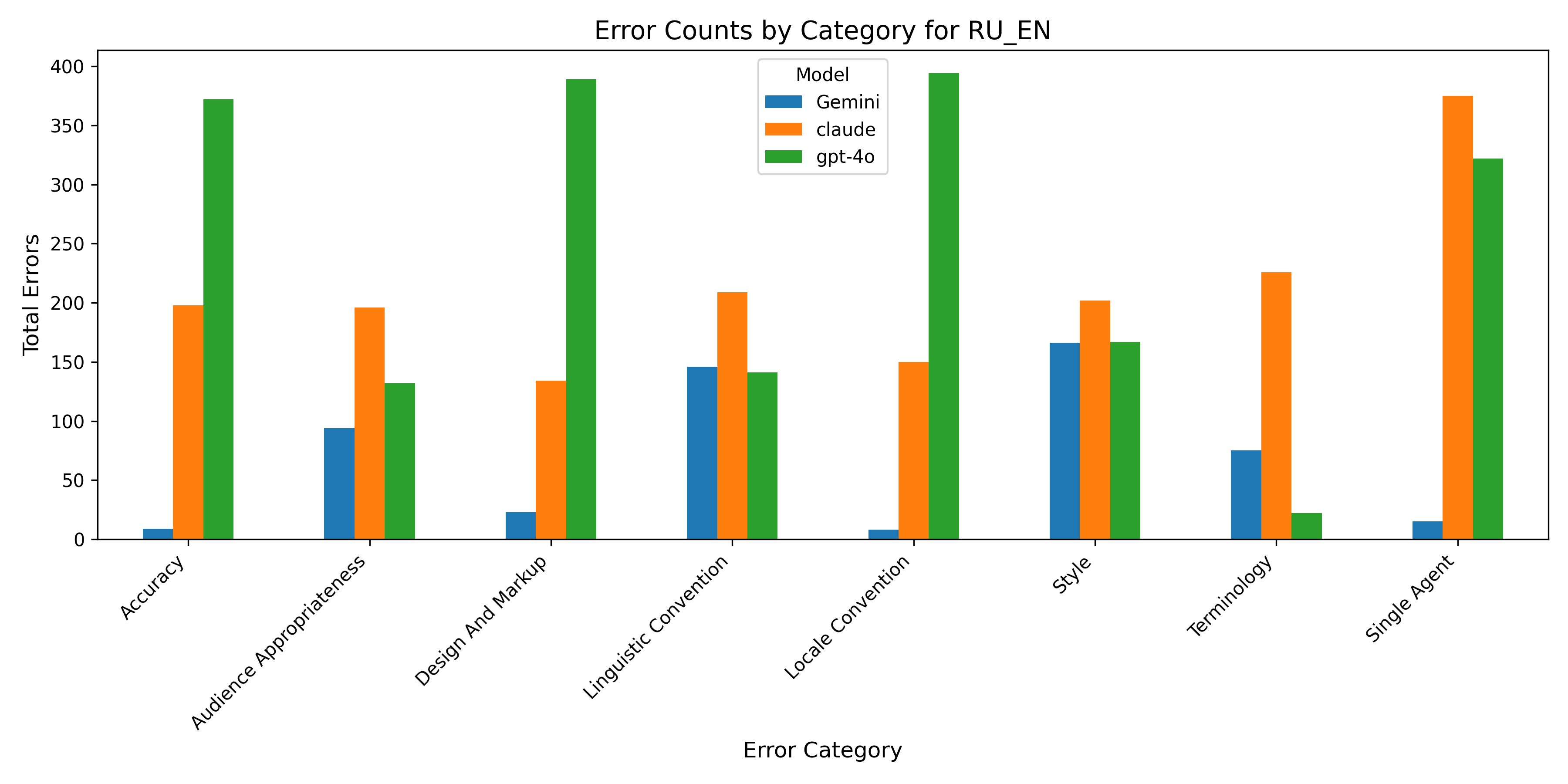}
        \caption*{RU$\rightarrow$EN}
    \end{minipage}

    \caption{Error count visualizations for each language pair. Left and right plots in each row represent forward and reverse translation directions.}
    \label{fig:appendix:error-by-language}
\end{figure*}

\clearpage
\onecolumn

\begin{figure}[htbp!]
    \centering
    \includegraphics[width=\linewidth]{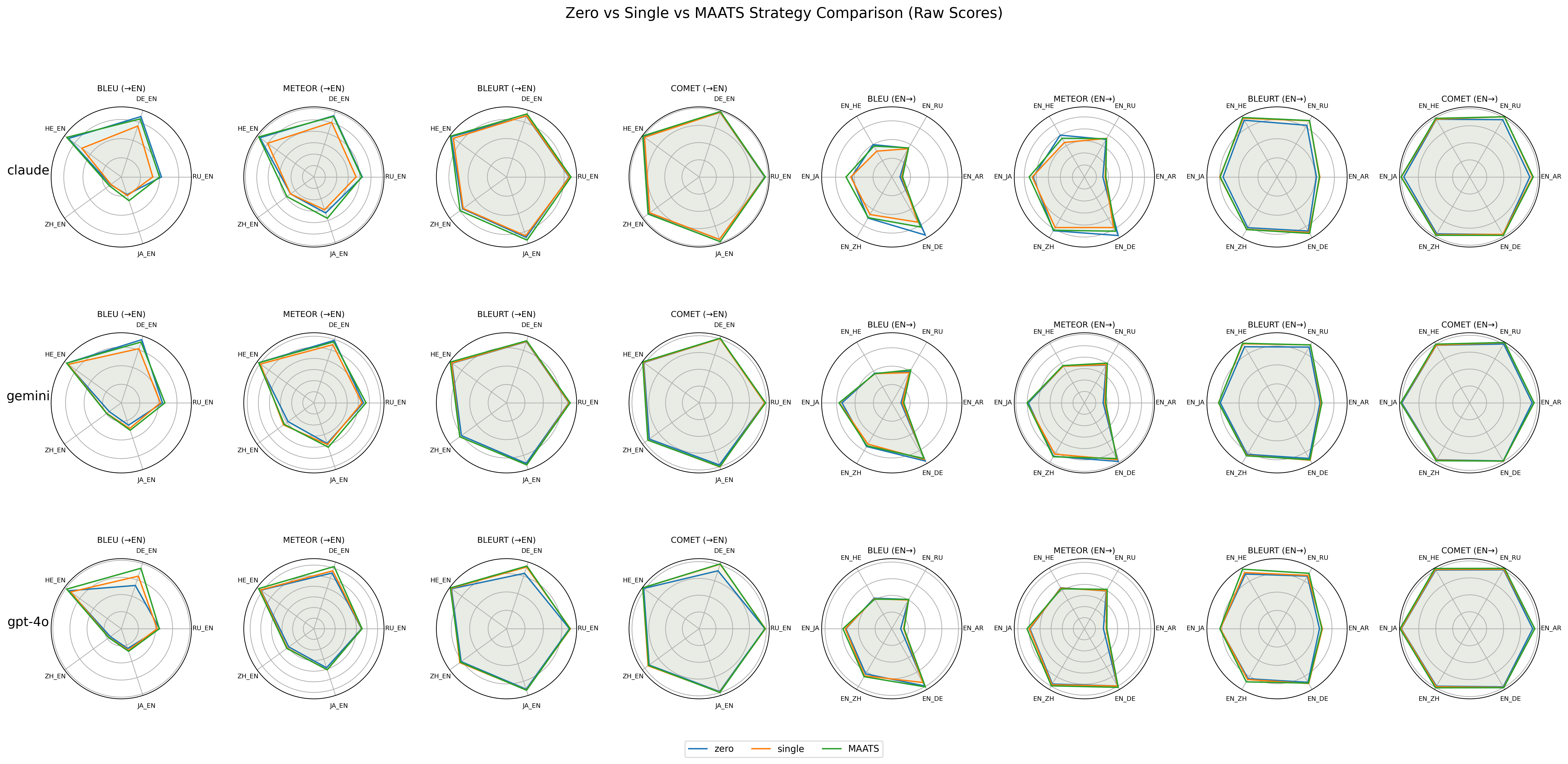}
    \caption{Full-Scale Radar Performance Chart for GPT-4o, Claude-3-haiku and Gemini-2.0-flash} 
    \label{fig:full scale}
\end{figure}

\begin{table*}[htbp]
\centering
\small
\renewcommand{\arraystretch}{1.2}
\setlength{\tabcolsep}{5pt}
\begin{tabular}{l|rrr|rrr|rrr}
\toprule
\textbf{Category} & \multicolumn{3}{c|}{\textbf{Claude}} & \multicolumn{3}{c|}{\textbf{GPT}} & \multicolumn{3}{c}{\textbf{Gemini}} \\
 & \textbf{TP} & \textbf{FP} & \textbf{FN} & \textbf{TP} & \textbf{FP} & \textbf{FN} & \textbf{TP} & \textbf{FP} & \textbf{FN} \\
\midrule
\multicolumn{10}{l}{\textbf{Single Agent}} \\
Accuracy          & 775 & 378  & 1144 & 795 & 654  & 1124 & 635 & 165  & 1284 \\
Fluency           & 150 & 242  & 721  & 138 & 212  & 733  & 153 & 306  & 718  \\
Locale Convention & 0   & 18   & 42   & 0   & 29   & 42   & 2   & 285  & 40   \\
Style             & 413 & 387  & 580  & 462 & 598  & 531  & 459 & 559  & 534  \\
Terminology       & 12  & 158  & 76   & 22  & 270  & 66   & 38  & 626  & 50   \\
\midrule
\multicolumn{10}{l}{\textbf{MAATS}} \\
Accuracy          & 942 & 462  & 977  & 1190 & 1533 & 519  & 752 & 503  & 1167 \\
Fluency           & 625 & 1213 & 246  & 506  & 964  & 249  & 620 & 1289 & 251  \\
Locale Convention & 28  & 1319 & 14   & 34   & 2681 & 7    & 15  & 531  & 27   \\
Style             & 735 & 976  & 258  & 646  & 855  & 247  & 731 & 976  & 262  \\
Terminology       & 76  & 1616 & 12   & 67   & 883  & 16   & 57  & 823  & 31   \\
\bottomrule
\end{tabular}
\caption{Confusion Matrix by Category and System Type: Claude, GPT, and Gemini – Single Agent vs. MAATS}
\label{tab:combined-confusion-matrix}
\end{table*}

\begin{table}[h!]
\resizebox{\columnwidth}{!}{%
\begin{tabular}{|l|l|c|c|c|c|c|}
\hline
\textbf{Language Pair} & \textbf{Model} & \textbf{BLEU} & \textbf{METEOR} & \textbf{BLEURT} & \textbf{COMET} & \textbf{Summary} \\
\hline
\multirow{3}{*}{EN\_RU} 
  & Claude   & \ding{51} & \ding{55} & \ding{51} & \ding{55} & 2/4 \\
  & Gemini   & \ding{51} & \ding{51} & \ding{55} & \ding{51} & 3/4 \\
  & GPT-4o   & \ding{51} & \ding{51} & \ding{51} & \ding{51} & 4/4 \\
\hline
\multirow{3}{*}{RU\_EN} 
  & Claude   & \ding{55} & \ding{55} & \ding{55} & \ding{51} & 1/4 \\
  & Gemini   & \ding{51} & \ding{51} & \ding{51} & \ding{51} & 4/4 \\
  & GPT-4o   & \ding{55} & \ding{51} & \ding{55} & \ding{55} & 1/4 \\
\hline
\multirow{3}{*}{EN\_DE} 
  & Claude   & \ding{55} & \ding{55} & \ding{51} & \ding{51} & 2/4 \\
  & Gemini   & \ding{55} & \ding{55} & \ding{55} & \ding{55} & 0/4 \\
  & GPT-4o   & \ding{51} & \ding{51} & \ding{55} & \ding{51} & 3/4 \\
\hline
\multirow{3}{*}{DE\_EN} 
  & Claude   & \ding{55} & \ding{55} & \ding{55} & \ding{51} & 1/4 \\
  & Gemini   & \ding{55} & \ding{55} & \ding{51} & \ding{51} & 2/4 \\
  & GPT-4o   & \ding{51} & \ding{51} & \ding{51} & \ding{51} & 4/4 \\
\hline
\multirow{3}{*}{EN\_HE} 
  & Claude   & \ding{55} & \ding{55} & \ding{51} & \ding{51} & 2/4 \\
  & Gemini   & \ding{55} & \ding{51} & \ding{51} & \ding{51} & 3/4 \\
  & GPT-4o   & \ding{55} & \ding{55} & \ding{51} & \ding{51} & 2/4 \\
\hline
\multirow{3}{*}{HE\_EN} 
  & Claude   & \ding{51} & \ding{51} & \ding{55} & \ding{51} & 3/4 \\
  & Gemini   & \ding{55} & \ding{55} & \ding{51} & \ding{51} & 2/4 \\
  & GPT-4o   & \ding{51} & \ding{51} & \ding{51} & \ding{51} & 4/4 \\
\hline
\multirow{3}{*}{EN\_JA} 
  & Claude   & \ding{51} & \ding{51} & \ding{55} & \ding{51} & 3/4 \\
  & Gemini   & \ding{51} & \ding{55} & \ding{55} & \ding{51} & 2/4 \\
  & GPT-4o   & \ding{51} & \ding{51} & \ding{51} & \ding{51} & 4/4 \\
\hline
\multirow{3}{*}{JA\_EN} 
  & Claude   & \ding{51} & \ding{51} & \ding{55} & \ding{51} & 3/4 \\
  & Gemini   & \ding{51} & \ding{51} & \ding{55} & \ding{55} & 2/4 \\
  & GPT-4o   & \ding{51} & \ding{55} & \ding{51} & \ding{51} & 3/4 \\
\hline
\multirow{3}{*}{EN\_ZH} 
  & Claude   & \ding{55} & \ding{55} & \ding{51} & \ding{51} & 2/4 \\
  & Gemini   & \ding{55} & \ding{51} & \ding{51} & \ding{51} & 3/4 \\
  & GPT-4o   & \ding{51} & \ding{51} & \ding{51} & \ding{51} & 4/4 \\
\hline
\multirow{3}{*}{ZH\_EN} 
  & Claude   & \ding{51} & \ding{51} & \ding{55} & \ding{51} & 3/4 \\
  & Gemini   & \ding{51} & \ding{55} & \ding{51} & \ding{51} & 3/4 \\
  & GPT-4o   & \ding{51} & \ding{51} & \ding{55} & \ding{55} & 2/4 \\
\hline
\multirow{3}{*}{EN\_AR} 
  & Claude   & \ding{51} & \ding{51} & \ding{55} & \ding{51} & 3/4 \\
  & Gemini   & \ding{51} & \ding{51} & \ding{51} & \ding{55} & 3/4 \\
  & GPT-4o   & \ding{51} & \ding{51} & \ding{55} & \ding{55} & 2/4 \\
\hline
\end{tabular}
}
\caption{Translation Gains (MAATS > Single-Agent > Zero-Shot) Across Language Pairs and Models}
\label{tab:pairwise}
\end{table}

\clearpage

\begin{table}[htbp]
\centering

\renewcommand{\arraystretch}{1.5}
\begin{CJK*}{UTF8}{gbsn} 
\begin{tabularx}{\textwidth}{>{\hsize=0.15\hsize}X >{\hsize=0.425\hsize}X >{\hsize=0.425\hsize}X}
\toprule
\textbf{Item} & \textbf{Single Agent} & \textbf{Multi-Agent} \\
\midrule
\textbf{Source Sentence} & I was looking for a good first car guy car; and I really love the 80s aesthetic. & Same as Single Agent \\
\textbf{Initial Translation} & 我在寻找一辆好的第一辆车，而且我真的很喜欢80年代的美学。 & Same as Single Agent \\
\midrule
\textbf{Accuracy} & 
\textbf{[Major]} Mistranslation – Failed to convey "car guy car" as a car for enthusiasts. & 
\textbf{[Critical]} Omission – "Car guy" omitted. \newline
\textbf{[Major]} Mistranslation – “第一辆车” does not convey enthusiast meaning. \\
\midrule
\textbf{Fluency} & 
\textbf{[Minor]} Punctuation – Semicolon translated as comma. & 
\textbf{[Minor]} Punctuation – Semicolon should be a comma in Chinese (“车；而且”). \\
\midrule
\textbf{Style} & 
\textbf{[Major]} Awkward – Literal rendering of "car guy car" is unnatural. & 
\textbf{[Minor]} Unnatural Flow – “好的第一辆车” sounds awkward. \newline
\textbf{[Minor]} Unnatural Flow – “80年代的美学” could be “80年代风格”. \\
\midrule
\textbf{Audience Appropriateness} & — & 
\textbf{[Major]} Culture-specific Reference – "Car guy car" may not be understood by Chinese readers; adaptation needed. \\
\midrule
\textbf{Terminology} & — & 
\textbf{[Critical]} Term Not Applied – "Car guy" was not translated per glossary. \\
\midrule
\textbf{Locale Conventions} & — & 
\textbf{[Minor]} Number Format – “第一辆车” doesn’t reflect enthusiast nuance. \\
\midrule
\textbf{Other Categories} & No issues marked in: Linguistic Conventions (other than punctuation), Non-Translation, and Other. & Same as left. \\
\midrule
\textbf{Final Translation} & 我在寻找一辆适合车迷的第一辆车；而且我真的很喜欢80年代的美学。 & 我在寻找一辆适合车迷的新手车，而且我真的很喜欢80年代风格。 \\
\bottomrule
\end{tabularx}
\end{CJK*} 
\caption{\textbf{Case Study 1}- Detailed MQM Evaluation of a Challenging Phrase: "Car Guy Car"}
\label{case study 1}
\end{table}

\clearpage
\begin{table}[htbp]
\centering

\renewcommand{\arraystretch}{1.5}
\begin{CJK*}{UTF8}{gbsn} 
\begin{tabularx}{\textwidth}{>{\hsize=0.15\hsize}X >{\hsize=0.425\hsize}X >{\hsize=0.425\hsize}X}
\toprule
\textbf{Item} & \textbf{Single Agent} & \textbf{Multi-Agent} \\
\midrule
\textbf{Source Sentence} & We provide links to where you can book the fare. & Same as Single Agent \\
\textbf{Initial Translation} & 我们提供可以预订票价的链接。 & Same as Single Agent \\
\midrule
\textbf{Accuracy} & 
\textbf{[Major]} Omission – The phrase “to where” is omitted, which specifies the destination or platform. & 
\textbf{[Critical]} Omission – “to where” is omitted from the translation. \newline
\textbf{[Major]} Mistranslation – “预订票价” is literal but doesn’t capture the contextual meaning of providing links for booking. \\
\midrule
\textbf{Fluency} & 
\textbf{[Minor]} Punctuation – Missing space after the period in the source. & 
\textbf{[Minor]} Punctuation – Missing period at the end of the Chinese sentence. \\
\midrule
\textbf{Style} & — & 
\textbf{[Minor]} Unnatural Flow – “可以预订票价的链接” is correct but slightly awkward. More natural: “可以预订机票的链接” or “票价预订链接”. \\
\midrule
\textbf{Audience Appropriateness} & — & 
\textbf{[Minor]} Culture-specific Reference – “预订票价” is not natural in Chinese. Suggested alternatives: “预订机票” or “预订航班”. \\
\midrule
\textbf{Terminology} & — & 
\textbf{[Critical]} Term Not Applied – “Fare” should be highlighted per glossary but wasn’t. \\
\midrule
\textbf{Locale Conventions} & — & No error. \\
\midrule
\textbf{Other Categories} & No issues marked in: Linguistic Conventions (other than punctuation), Non-Translation, and Other. & Same as left. \\
\midrule
\textbf{Final Translation} & 我们提供链接，您可以通过这些链接预订票价。 & 我们提供可以预订机票的链接。 \\
\bottomrule
\end{tabularx}
\end{CJK*} 
\caption{\textbf{Case Study 2} – MQM-Based Comparative Analysis of Booking-Related Translation: “Book the Fare”}
\label{case study 2}
\end{table}
\onecolumn

\clearpage
\begin{table}[htbp]
\centering
\centering

\renewcommand{\arraystretch}{1.5}
\begin{CJK*}{UTF8}{gbsn} 
\begin{tabularx}{\textwidth}{>{\hsize=0.15\hsize}X >{\hsize=0.425\hsize}X >{\hsize=0.425\hsize}X}
\toprule
\textbf{Item} & \textbf{Single Agent} & \textbf{Multi-Agent} \\
\midrule
\textbf{Source Sentence} & I started a new shawl. & Same as Single Agent \\
\textbf{Initial Translation} & 我开始了一条新的披肩。 & Same as Single Agent \\
\midrule
\textbf{Accuracy} & — & 
\textbf{[Minor]} Addition – The word “一条” was added in the translation. While it is not present in the source, it does not significantly affect meaning. \\
\midrule
\textbf{Fluency} & — & No error. \\
\midrule
\textbf{Style} & — & 
\textbf{[Minor]} Unnatural Flow – “我开始了一条新的披肩” is grammatically correct, but a more natural Chinese expression is “我开始织一条新的披肩”. \\
\midrule
\textbf{Audience Appropriateness} & — & 
\textbf{[Minor]} Culture-specific Reference – “披肩” is appropriate and understandable; no adaptation needed. \\
\midrule
\textbf{Terminology} & — & No error. \\
\midrule
\textbf{Locale Conventions} & — & No error. \\
\midrule
\textbf{Other Categories} & No issues marked in any other dimensions. & Same as left. \\
\midrule
\textbf{Final Translation} & 我开始了一条新的披肩。 & 我开始织一条新的披肩。 \\
\bottomrule
\end{tabularx}
\end{CJK*} 
\caption{\textbf{Case Study 3.1} – Subtle Stylistic Refinement in “I Started a New Shawl”}
\label{case study 3.1}
\end{table}
\begin{table}[htbp]

\centering

\renewcommand{\arraystretch}{1.5}
\begin{CJK*}{UTF8}{gbsn} 
\begin{tabularx}{\textwidth}{>{\hsize=0.15\hsize}X >{\hsize=0.425\hsize}X >{\hsize=0.425\hsize}X}
\toprule
\textbf{Item} & \textbf{Single Agent} & \textbf{Multi-Agent} \\
\midrule
\textbf{Source Sentence} & The literature class hatred is even worse: I love to read. & Same as Single Agent \\
\textbf{Initial Translation} & 文学课的仇恨更严重：我喜欢阅读。 & Same as Single Agent \\
\midrule
\textbf{Accuracy} & 
\textbf{[Major]} Mistranslation – The phrase “The literature class hatred” is misinterpreted and suggests something more severe than intended. & 
\textbf{[Major]} Mistranslation – “仇恨” correctly means “hatred,” but fails to reflect the tone/context of dislike within an academic subject. \\
\midrule
\textbf{Fluency} & 
\textbf{[Minor]} Punctuation – The colon “：” may not be the best fit for Chinese sentence structure. & 
\textbf{[Minor]} Punctuation – The colon should be a full-width colon “：” to match Chinese conventions. \\
\midrule
\textbf{Style} & 
\textbf{[Major]} Awkward – The structure is grammatically acceptable but feels unnatural in Chinese. & 
\textbf{[Major]} Unnatural Flow – “文学课的仇恨更严重” is awkward; better options include “我对文学课的厌恶更深” or “反感更强”. \\
\midrule
\textbf{Audience Appropriateness} & — & 
\textbf{[Major]} Culture-specific Reference – The phrase “class hatred” may not culturally align with Chinese interpretation of classroom sentiments. \\
\midrule
\textbf{Terminology} & — & No error. \\
\midrule
\textbf{Locale Conventions} & — & No error. \\
\midrule
\textbf{Other Categories} & No issues marked in any other dimensions. & Same as left. \\
\midrule
\textbf{Final Translation} & 文学课上的仇恨更糟糕，我却喜欢阅读。 & 文学课的厌恶更深：我喜欢阅读。 \\
\bottomrule
\end{tabularx}
\end{CJK*} 
\caption{\textbf{Case Study 3.2} – MQM Assessment of Emotional Tone and Cultural Interpretation in “The Literature Class Hatred”}
\label{case study 3.2}
\end{table}

\onecolumn
\setstretch{1.05}
\section{Single-Agent Annotation Prompt}
\label{single-agent prompt}
\begin{tcolorbox}[colback=white, colframe=black, boxrule=0.4pt, sharp corners, width=\textwidth]

You are a professional annotator responsible for evaluating the quality of machine translation output based on the Multidimensional Quality Metrics (MQM) framework. \\
Your task is to identify all errors in the translation and classify them into the following dimensions: \\
\textbf{Accuracy} (Mistranslation, omission, addition, over-translation), \\
\textbf{Terminology} (Incorrect term choice, missing glossary application), \\
\textbf{Linguistic Conventions} (Grammar mistakes, punctuation errors, typographical issues), \\
\textbf{Style} (Unnatural flow, awkward expressions, overly literal phrasing), \\
\textbf{Locale Conventions} (Wrong format, incorrect symbols), \\
\textbf{Audience Appropriateness} (Overly technical terms for general audiences, offensive wording), \\
\textbf{Design and Markup} (Missing markup, broken text formatting, layout inconsistencies), \\
\noindent\rule{\textwidth}{0.4pt}

\textbf{Each error must be assigned a severity level:} \\
\textbf{Critical:} Critical errors inhibit comprehension of the text. \\
\textbf{Major:} Major errors disrupt the flow, but what the text is trying to say is still understandable. \\
\textbf{Minor:} Minor errors are technically errors, but do not disrupt the flow or hinder comprehension. \\
If no error is detected, return \texttt{"no-error"} in its severity.

\noindent\rule{\textwidth}{0.4pt}

\textbf{Your answer should follow the following template:}

\medskip
\texttt{MQM annotations:} \\
\texttt{Accuracy Errors} \\
\texttt{[Critical]: [error/error\_subcategory] - [brief explanation]} \\
\texttt{[Major]: [error/error\_subcategory] - [brief explanation]} \\
\texttt{[Minor]: [error/error\_subcategory] - [brief explanation]} \\

\texttt{Terminology Errors} \\
\texttt{[Critical]: [error/error\_subcategory] - [brief explanation]} \\
\texttt{[Major]: [error/error\_subcategory] - [brief explanation]} \\
\texttt{[Minor]: [error/error\_subcategory] - [brief explanation]} \\



.....

\texttt{Audience Appropriateness Errors} \\
\texttt{[Critical]: [error/error\_subcategory] - [brief explanation]} \\
\texttt{[Major]: [error/error\_subcategory] - [brief explanation]} \\
\texttt{[Minor]: [error/error\_subcategory] - [brief explanation]} \\

\texttt{Design and Markup Errors} \\
\texttt{[Critical]: [error/error\_subcategory] - [brief explanation]} \\
\texttt{[Major]: [error/error\_subcategory] - [brief explanation]} \\
\texttt{[Minor]: [error/error\_subcategory] - [brief explanation]} \\

\medskip
Please follow the template, put \texttt{[Error Severity Levels]:[error/error\_subcategory] - None if doesn't exit}

\end{tcolorbox}


\clearpage
\onecolumn
\setstretch{1.05}  
\section{MAATS Evaluator Prompts}

\section*{Accuracy Annotation Prompt}
\label{AccuracyPrompt}
\begin{tcolorbox}[colback=white, colframe=black, boxrule=0.4pt, sharp corners, width=\textwidth]

You are responsible for evaluating the accuracy of machine-translated content. Your main task is to identify errors in the translation and assess their severity. Focus specifically on \textbf{accuracy-related problems}—these arise when:

\textbf{–} The translated content fails to capture the exact meaning intended in the source. \\
\textbf{–} The translation, even if correct in isolation, does not align properly with the surrounding context.

\textbf{Accuracy errors are categorized as follows:} \\
\textbf{1. Addition:} Extra words are added that are not in the source and do not improve meaning. \\
\textbf{2. Mistranslation:} The meaning is incorrectly conveyed due to wrong word choice or unnatural phrasing, even if grammatically fine. \\
\textbf{3. MT Hallucination:} The translation is fluent but entirely unrelated to the source (e.g., invented or repeated content). \\
\textbf{4. Omission:} Important content from the source is missing, affecting meaning. \\
\textbf{5. Untranslated:} Source text appears in the target without translation (except named entities). \\
\textbf{6. Wrong Named Entity:} Errors in proper names or entities (e.g., spelling, translation, or form issues).

\noindent\rule{\textwidth}{0.4pt}

\textbf{Error Severity Levels:} \\
\textbf{– Critical:} Inhibits comprehension of the text. \\
\textbf{– Major:} Disrupts the flow of the text but the intended meaning remains understandable. \\
\textbf{– Minor:} Technical errors that do not significantly hinder comprehension.

\noindent\rule{\textwidth}{0.4pt}

\textbf{Example} \\
\textbf{Source (English):} That way you can be sure that you were the one who made the changes. \\
\textbf{Target (Spanish):} Así puedes estar seguro de que fuiste tú quien hizo todos los cambios. \\
\textbf{MQM annotations:} \\
\textbf{major: accuracy/addition} – “Todos” (“all”) is not present in the source and has been unnecessarily added, altering the original meaning.

\noindent\rule{\textwidth}{0.4pt}

\textbf{Output Format} \\
Learn from the example, based on the source segment and the machine translation provided, identify the error types in the translation and classify them using the following template:

\medskip
\texttt{MQM annotations:} \\
\texttt{Accuracy Errors} \\
\texttt{[Critical]: [accuracy/error\_subcategory] - Brief explanation.} \\
\texttt{[Major]: [accuracy/error\_subcategory] - Brief explanation.} \\
\texttt{[Minor]: [accuracy/error\_subcategory] - Brief explanation.}

\medskip
Please follow the template, put 

\texttt{[Error Severity Levels]:[audience\_appropriateness/error\_subcategory] - None}

If doesn't exist
\end{tcolorbox}

\twocolumn


\clearpage
\onecolumn
\setstretch{1.05}
\section*{Audience Appropriateness Annotation Prompt}
\label{AudiencePrompt}

\begin{tcolorbox}[colback=white, colframe=black, boxrule=0.4pt, sharp corners, width=\textwidth]

You are responsible for evaluating the audience appropriateness of machine-translated content. Your main task is to detect audience appropriateness issues and judge how severely they affect the reader’s experience. \\
Focus specifically on Audience Appropriateness errors—these arise when: \\
\textbf{–} The translation includes content that feels out of place, confusing, or culturally insensitive for the target readers. \\
\textbf{–} The language variety or cultural references are misaligned with the intended audience, leading to misunderstanding or alienation.

\textbf{Audience Appropriateness errors are categorized as follows:} \\
\textbf{1. Culture-Specific Reference:} A metaphor, idiom, term, or expression is used that does not fit the audience’s cultural context, making the message unclear or inappropriate. \\
\textbf{2. Wrong Language Variety:} The translation uses a different regional dialect, spelling, or phrasing than what the audience expects (e.g., UK English instead of US English).

\noindent\rule{\textwidth}{0.4pt}

\textbf{Severity Levels} \\
\textbf{Critical:} Severely hinders understanding or creates major cultural offense or disconnect. \\
\textbf{Major:} Clearly disrupts the expected tone or context, causing noticeable confusion. \\
\textbf{Minor:} Slightly reduces the naturalness or relatability for the audience, though meaning is preserved.

\noindent\rule{\textwidth}{0.4pt}

\begin{CJK*}{UTF8}{min}
\textbf{Example} \\
\textbf{Source (English):} Please contact our support team if you experience any issues. \\
\textbf{Target (Japanese):} お客様ごとに担当者が異なりますので、まずは担当者にご相談ください \\
\textbf{MQM annotations:} \\
\textbf{major: audience\_appropriate/culture\_specific\_reference} – The phrase “お客様ごとに担当者が異なりますので…” (“each customer has a different representative”) assumes a Japan-specific corporate support structure that may not reflect the source content. It introduces a culturally specific reference not present in the source and could mislead or confuse international audiences.
\end{CJK*}

\noindent\rule{\textwidth}{0.4pt}

\textbf{Output Format} \\
Learn from the example, based on the source segment and the machine translation provided, identify the error types in the translation and classify them using the following template:

\medskip
\texttt{MQM annotations:} \\
\texttt{Audience Appropriateness Errors} \\
\texttt{[Critical]: [audience\_appropriateness/error\_subcategory] - Brief explanation.} \\
\texttt{[Major]: [audience\_appropriateness/error\_subcategory] - Brief explanation.} \\
\texttt{[Minor]: [audience\_appropriateness/error\_subcategory] - Brief explanation.}

\medskip
Please follow the template, put 

\texttt{[Error Severity Levels]:[audience\_appropriateness/error\_subcategory] - None}

If doesn't exit

\end{tcolorbox}

\vspace{3em}



\setstretch{1.05}
\section*{Design and Markup Annotation Prompt}
\label{DesignPrompt}

\begin{tcolorbox}[colback=white, colframe=black, boxrule=0.4pt, sharp corners, width=\textwidth]

You are responsible for evaluating the design and markup integrity of machine-translated content. Your main task is to detect design or formatting issues and judge how severely they impact the clarity, layout, or usability of the translation. \\
Focus specifically on Design and Markup errors—these arise when: \\
\textbf{–} The translation introduces problems with HTML or XML tags, character encoding, or other structural elements. \\
\textbf{–} Visual elements such as emoji formatting are altered, malformed, or broken during translation.

\textbf{Design and Markup errors are categorized as follows:} \\
\textbf{Markup Tag:} Occurs when there are incorrect or malformed markup tags or components. This includes: \\
\textbf{–} Improperly escaped HTML entities (e.g., \& \#160; instead of \#160;); \\
\textbf{–} Inconsistent or broken emoji representations; \\
\textbf{–} Extra spaces inserted within HTML symbols or code.

\noindent\rule{\textwidth}{0.4pt}

\textbf{Severity Levels} \\
\textbf{Critical:} Severely affects readability, layout, or breaks content structure (e.g., broken HTML tags that cause rendering failures). \\
\textbf{Major:} Disrupts the visual flow or formatting but leaves content mostly understandable. \\
\textbf{Minor:} Small formatting glitches that do not significantly impair readability or function.

\noindent\rule{\textwidth}{0.4pt}
\begin{CJK*}{UTF8}{gbsn}
\textbf{Example} \\
\textbf{Source (Chinese):} 请点击<了解更多>以查看更多信息。 \\
\end{CJK*}
\textbf{Target (English):} Please click \&lt;Learn More\&gt; to view more information. \\
\textbf{MQM annotations:} \\
\textbf{critical: design\_and\_markup/markup\_tag} – “\&lt;” and “\&gt;” contain extra spaces and are malformed HTML entities. \\
\textbf{major: design\_and\_markup/markup\_tag} – Angle brackets are incorrectly displayed as literal text, which may confuse readers.

\noindent\rule{\textwidth}{0.4pt}

\textbf{Output Format} \\
Learn from the example, based on the source segment and the machine translation provided, identify the error types in the translation and classify them using the following template:

\medskip
\texttt{MQM annotations:} \\
\texttt{Design and Markup Errors} \\
\texttt{[Critical]: [design\_and\_markup/error\_subcategory] - Brief explanation.} \\
\texttt{[Major]: [design\_and\_markup/error\_subcategory] - Brief explanation.} \\
\texttt{[Minor]: [design\_and\_markup/error\_subcategory] - Brief explanation.}

\medskip
Please follow the template, put 

\texttt{[Error Severity Levels]:[design\_and\_markup/error\_subcategory] - None}

If doesn't exit

\end{tcolorbox}

\vspace{5em}

\setstretch{1.05}
\section*{Linguistic Conventions Annotation Prompt}
\tcbset{before skip=6pt, after skip=6pt, boxsep=3pt, left=3pt, right=3pt, top=3pt, bottom=3pt}
\label{LinguisticPrompt}
\begin{tcolorbox}[colback=white, colframe=black, boxrule=0.4pt, sharp corners, width=\textwidth]

You are responsible for evaluating the linguistic correctness of machine-translated content. Your task is to identify issues related to standard writing rules in the target language, regardless of whether the text is a translation. \\
Focus specifically on Linguistic Conventions errors—these arise when: \\
\textbf{–} The target text violates standard rules of grammar, spelling, punctuation, or formatting. \\
\textbf{–} The translation is linguistically awkward or incorrect, even if the meaning is understandable.

\textbf{Linguistic Conventions errors are categorized as follows:} \\
\textbf{1. Agreement:} Issues where words do not agree in gender, number, case, or person (e.g., “they was” instead of “they were”). \\
\textbf{2. Capitalization:} Incorrect use of upper- or lowercase letters (e.g., “internet” vs. “Internet,” or “hello” at the start of a sentence). \\
\textbf{3. Grammar:} Problems with morphology or syntax such as verb tense, word form, or function words (e.g., “He go to school” instead of “He goes to school”). \\
\textbf{4. Punctuation:} Misuse or omission of punctuation marks, including missing closing quotation marks, incorrect sentence endings, or replacing a colon with a comma. \\
\textbf{5. Spelling:} Misspelled words, missing accents or diacritics, or incorrect hyphenation within a word (e.g., “co-operate” vs. “cooperate”). \\
\textbf{6. Whitespace:} Extra or missing spaces between words or characters (e.g., “in credible” instead of “incredible”; “nextto” instead of “next to”). \\
\textbf{7. Word Order:} Words appear in an unnatural or incorrect sequence, affecting sentence flow or clarity.

\noindent\rule{\textwidth}{0.4pt}

\textbf{Severity Levels} \\
\textbf{Critical:} Severely disrupts comprehension; the sentence may become unreadable or misleading. \\
\textbf{Major:} Breaks the grammatical flow or makes reading difficult, but the general meaning is still clear. \\
\textbf{Minor:} Small errors (e.g., typos or punctuation slips) that do not significantly affect understanding.

\noindent\rule{\textwidth}{0.4pt}

\textbf{Example} \\
\textbf{Source (English):} With regards to your query, I would like to inform you that your device is a Wi-Fi device. \\
\textbf{Target (Spanish):} Con respecto a su consulta, me gustaría informarle de que su dispositivo es un dispositivo Wi-Fi. \\
\textbf{MQM annotations:} \\
\textbf{minor: linguistic\_convention/spelling} – “Wi-Fi” includes an unnecessary hyphen in Spanish; the correct spelling is “Wifi” without the hyphen.

\noindent\rule{\textwidth}{0.4pt}

\textbf{Output Format} \\
Learn from the example, based on the source segment and the machine translation provided, identify the error types in the translation and classify them using the following template:

\medskip
\texttt{MQM annotations:} \\
\texttt{Linguistic Conventions Errors} \\
\texttt{[Critical]: [linguistic\_conventions/error\_subcategory] - Brief explanation.} \\
\texttt{[Major]: [linguistic\_conventions/error\_subcategory] - Brief explanation.} \\
\texttt{[Minor]: [linguistic\_conventions/error\_subcategory] - Brief explanation.}

\medskip
Please follow the template, put 

\texttt{[Error Severity Levels]:[linguistic\_conventions/error\_subcategory] - None}

if doesn't exit

\end{tcolorbox}


\setstretch{1.05}
\section*{Locale Conventions Annotation Prompt}
\label{LocalePrompt}
\begin{tcolorbox}[colback=white, colframe=black, boxrule=0.4pt, sharp corners, width=\textwidth]

You are responsible for evaluating the locale-specific formatting and conventions of machine-translated content. Your task is to identify issues where the translation fails to follow regional standards or client-specific formatting requirements. \\
Focus specifically on Locale Conventions errors—these arise when: \\
\textbf{–} The translation includes content that does not match regional formats for things like dates, numbers, addresses, or telephone numbers. \\
\textbf{–} Client instructions for localized formatting are ignored, even if the text is grammatically correct.

\textbf{Locale Conventions errors are categorized as follows:} \\
\textbf{1. Address Format:} The structure of an address does not follow the local convention (e.g., writing city before postal code in a region where the reverse is standard). \\
\textbf{2. Currency Format:} Currency symbols, abbreviations, or placement are incorrect (e.g., using “\$100” instead of “100€” for European locales). \\
\textbf{3. Date/Time Format:} Date or time expressions do not align with the regional norm (e.g., “03/04/2023” for UK should mean 3 April, not March 4). \\
\textbf{4. Measurement Format:} Use of measurement units (e.g., inches, meters, grams) that are inappropriate for the locale or formatted incorrectly (e.g., “5ft” instead of “1.52m”). \\
\textbf{5. Number Format:} Digits, separators, or groupings deviate from locale rules (e.g., “1,000.00” in US vs. “1.000,00” in many EU countries). \\
\textbf{6. Telephone Format:} Phone numbers do not follow regional presentation (e.g., missing country code or incorrect spacing).

\noindent\rule{\textwidth}{0.4pt}
\begin{CJK*}{UTF8}{gbsn}
\textbf{Example} \\
\textbf{Source (Chinese):} 活动时间：2023年4月3日 \\
\end{CJK*}
\textbf{Target (English):} Event date: 04/03/2023 \\
\textbf{Locale:} United Kingdom \\
\textbf{MQM annotations:} \\
\textbf{critical: locale\_conventions/date\_time\_format} – “04/03/2023” could be read as April 3 instead of 3 April, which contradicts UK format and may cause misunderstanding.

\noindent\rule{\textwidth}{0.4pt}

\textbf{Output Format} \\
Learn from the example, based on the source segment and the machine translation provided, identify the error types in the translation and classify them using the following template:

\medskip
\texttt{MQM annotations:} \\
\texttt{Locale Conventions Errors} \\
\texttt{[Critical]: [locale\_conventions/error\_subcategory] - Brief explanation.} \\
\texttt{[Major]: [locale\_conventions/error\_subcategory] - Brief explanation.} \\
\texttt{[Minor]: [locale\_conventions/error\_subcategory] - Brief explanation.}

\medskip
Please follow the template, put 

\texttt{[Error Severity Levels]:[locale\_conventions/error\_subcategory] - None}

if doesn't exit

\end{tcolorbox}


\setstretch{1.05}
\section*{Style Annotation Prompt}
\begin{tcolorbox}[colback=white, colframe=black, boxrule=0.4pt, sharp corners, width=\textwidth]

You are responsible for evaluating the stylistic quality of machine-translated content. Your task is to identify issues related to fluency, tone, naturalness, and adherence to client style preferences. \\
Focus specifically on Style errors—these arise when: \\
\textbf{–} The translation reads unnaturally, awkwardly, or too literally. \\
\textbf{–} The output ignores client-specific tone, register, or stylistic guidelines, even if the meaning is accurate.

\textbf{Style errors are categorized as follows:} \\
\textbf{1. Company Style:} The translation fails to comply with company or client-specific guidelines (e.g., using passive voice where active is required). \\
\textbf{2. Do Not Translate:} A phrase or brand name was translated even though it should have been left in the original language, per client instructions. \\
\textbf{3. Inconsistency:} Key terms, expressions, or stylistic choices are not used consistently throughout the content (e.g., switching between “Sign in” and “Log in”). \\
\textbf{4. Lacks Creativity:} The translation is correct but lacks variation, nuance, or marketing appeal expected by the client, especially in creative or promotional content. \\
\textbf{5. Register:} The formality level is inappropriate for the context (e.g., too casual in a legal document or too formal in a gaming app). \\
\textbf{6. Unnatural Flow:} The translation sounds robotic, stilted, or too close to the source structure, making it awkward in the target language.

\noindent\rule{\textwidth}{0.4pt}

\textbf{Severity Levels} \\
\textbf{Critical:} Severely affects readability or makes the tone completely inappropriate or confusing. \\
\textbf{Major:} Clearly disrupts the flow or tone, reducing clarity or engagement. \\
\textbf{Minor:} Slight awkwardness or tone mismatch that doesn’t block understanding.

\noindent\rule{\textwidth}{0.4pt}

\textbf{Example} \\
\textbf{Source (English):} Click the gear icon; Click on Save \\
\textbf{Target (French):} Cliquez sur l’icône de l’engrenage; Cliquer sur Enregistrer \\
\textbf{MQM annotations:} \\
\textbf{major: style/inconsistency} – Mixed use of imperative (Cliquez) and infinitive (Cliquer) forms creates an inconsistency in tone. Although either form is acceptable in isolation, both should not appear together in the same instructional context.

\noindent\rule{\textwidth}{0.4pt}

\textbf{Output Format} \\
Learn from the example, based on the source segment and the machine translation provided, identify the error types in the translation and classify them using the following template:

\medskip
\texttt{MQM annotations:} \\
\texttt{Style Errors} \\
\texttt{[style]: [style/error\_subcategory] - Brief explanation.} \\
\texttt{[style]: [style/error\_subcategory] - Brief explanation.} \\
\texttt{[style]: [style/error\_subcategory] - Brief explanation.}

\medskip
Please follow the template, put \texttt{[Error Severity Levels]:[style/error\_subcategory] - None if doesn't exit}

\end{tcolorbox}


\setstretch{1.05}
\section*{Terminology Annotation Prompt}
\begin{tcolorbox}[colback=white, colframe=black, boxrule=0.4pt, sharp corners, width=\textwidth]

You are responsible for evaluating the correct use of glossary terms in machine-translated content. Your task is to identify issues where required terminology is not used correctly or fails to match the glossary specification. \\
Focus specifically on Terminology errors—these arise when: \\
\textbf{–} A glossary term (highlighted in blue in the Annotation Tool) is missing, misapplied, misspelled, or grammatically incorrect. \\
\textbf{–} The translation includes the correct glossary term but uses it in a contextually inappropriate way.

\textbf{Terminology errors are categorized as follows:} \\
\textbf{1. Term Not Applied:} A required glossary term was not used; a different or incorrect term was inserted instead, violating glossary rules. \\
\textbf{2. Wrong Term:} The glossary term appears but is used incorrectly in context—this includes typos, capitalization issues, plural/singular mismatches, or improper grammatical inflection.

\noindent\rule{\textwidth}{0.4pt}

\textbf{Severity Levels} \\
\textbf{Critical:} Glossary misuse leads to confusion, misinterpretation, or disrupts key meaning. \\
\textbf{Major:} The error affects tone or grammatical flow, though the meaning remains somewhat clear. \\
\textbf{Minor:} Cosmetic or minor issues such as case sensitivity or small typos that don’t impact comprehension.

\noindent\rule{\textwidth}{0.4pt}

\begin{CJK*}{UTF8}{gbsn}
\textbf{Example} \\
\textbf{Source (English):} S/N (Serial Number) \\
\end{CJK*}
\selectlanguage{russian}
\textbf{Target (Russian):} серийный номер (серийный номер) \\
\selectlanguage{english}
\textbf{MQM annotations:} \\
\textbf{major: terminology/wrong\_term –} \foreignlanguage{russian}{“серийный номер”} was used instead of the client-specified glossary entry \texttt{S/N}, leading to duplication of meaning and failure to follow the termbase.

\noindent\rule{\textwidth}{0.4pt}

\textbf{Output Format} \\
Learn from the example, based on the source segment and the machine translation provided, identify the error types in the translation and classify them using the following template:

\medskip
\texttt{MQM annotations:} \\
\texttt{Terminology Errors} \\
\texttt{[terminology]: [terminology/error\_subcategory] - Brief explanation.} \\
\texttt{[terminology]: [terminology/error\_subcategory] - Brief explanation.} \\
\texttt{[terminology]: [terminology/error\_subcategory] - Brief explanation.}

\medskip
Please follow the template, put \texttt{[Error Severity Levels]:[terminology/error\_subcategory] - None if doesn't exit}

\end{tcolorbox}

\newpage
\section{MAATS Editor Prompt}
\label{MAATS editor prompt}

\setstretch{1.05}

\begin{tcolorbox}[colback=white, colframe=black, boxrule=0.4pt, sharp corners, width=\textwidth]

You are an advanced translation refinement agent. Your task is to review all MQM-style annotation inputs across the following categories: Accuracy, Linguistic Conventions, Terminology, Style, Locale Conventions, Audience Appropriateness, and Design and Markup. \\
Your objective is to produce a revised translation that fully addresses and corrects all identified issues.


\textbf{Severity levels are defined as follows:} \\
\textbf{Critical:} Errors that prevent understanding or result in serious misinterpretation. \\
\textbf{Major:} Errors that disrupt the natural flow or clarity, but the overall meaning remains accessible. \\
\textbf{Minor:} Errors that are technically incorrect but do not significantly impact readability or comprehension.

\noindent\rule{\textwidth}{0.4pt}

\textbf{Apply the following correction priorities:} \\
Always resolve critical errors before addressing major or minor ones. \\
If a major correction conflicts with a minor suggestion, the major correction takes precedence. \\
In cases of conflict between critical and major/minor corrections, prioritize the critical fix. \\
When no conflict exists, incorporate both major and minor suggestions in a smooth and coherent manner.

\noindent\rule{\textwidth}{0.4pt}

\textbf{Input Format} \\
You will receive a set of annotations classified by error type and severity (e.g., major: accuracy/mistranslation, minor: style/inconsistency).

\textbf{Output Format} \\
Your output must be: \\
A single improved translation sentence that incorporates all necessary corrections. \\
Do not include explanations, labels, categories, or summaries—just the final revised sentence.

\end{tcolorbox}




\begin{thebibliography}{22}
\expandafter\ifx\csname natexlab\endcsname\relax\def\natexlab#1{#1}\fi

\bibitem[{Anthropic(2024)}]{anthropic2024claude3}
Anthropic. 2024.
\newblock \href {https://www.anthropic.com/news/claude-3-family} {The claude 3 model family: Opus, sonnet, haiku}.
\newblock Accessed: 2025-05-18.

\bibitem[{Banerjee and Lavie(2005)}]{banerjee-lavie-2005-meteor}
Satanjeev Banerjee and Alon Lavie. 2005.
\newblock \href {https://aclanthology.org/W05-0909/} {Meteor: An automatic metric for mt evaluation with improved correlation with human judgments}.
\newblock In \emph{Proceedings of the ACL Workshop on Intrinsic and Extrinsic Evaluation Measures for Machine Translation and/or Summarization}, pages 65--72, Ann Arbor, Michigan. Association for Computational Linguistics.

\bibitem[{DeepMind(2025)}]{google2025gemini}
Google DeepMind. 2025.
\newblock \href {https://cloud.google.com/vertex-ai/generative-ai/docs/models/gemini/2-0-flash} {Gemini 2.0 flash | generative ai on vertex ai}.
\newblock Accessed: 2025-05-18.

\bibitem[{Emerson(2013)}]{emerson2013original}
Peter Emerson. 2013.
\newblock \href {https://doi.org/10.1007/s00355-011-0603-9} {The original borda count and partial voting}.
\newblock \emph{Social Choice and Welfare}, 40(2):353--358.

\bibitem[{Feng et~al.(2024)Feng, Zhang, Li, Wu, Liao, Liu, Lang, Feng, Wu, and Liu}]{feng2024tearimprovingllmbasedmachine}
Zhaopeng Feng, Yan Zhang, Hao Li, Bei Wu, Jiayu Liao, Wenqiang Liu, Jun Lang, Yang Feng, Jian Wu, and Zuozhu Liu. 2024.
\newblock \href {http://arxiv.org/abs/2402.16379} {Tear: Improving llm-based machine translation with systematic self-refinement}.

\bibitem[{Freitag et~al.(2021{\natexlab{a}})Freitag, Foster, Grangier, Ratnakar, Tan, and Macherey}]{Freitag_2021}
Markus Freitag, George Foster, David Grangier, Viresh Ratnakar, Qijun Tan, and Wolfgang Macherey. 2021{\natexlab{a}}.
\newblock \href {https://doi.org/10.1162/tacl_a_00437} {Experts, errors, and context: A large-scale study of human evaluation for machine translation}.
\newblock \emph{Transactions of the Association for Computational Linguistics}, 9:1460–1474.

\bibitem[{Freitag et~al.(2021{\natexlab{b}})Freitag, Foster, Grangier, Ratnakar, Tan, and Macherey}]{freitag2021experts}
Markus Freitag, George Foster, David Grangier, Viresh Ratnakar, Qijun Tan, and Wolfgang Macherey. 2021{\natexlab{b}}.
\newblock \href {http://arxiv.org/abs/2104.14478} {Experts, errors, and context: A large-scale study of human evaluation for machine translation}.

\bibitem[{Jiao et~al.(2023)Jiao, Wang, tse Huang, Wang, Shi, and Tu}]{jiao2023chatgptgoodtranslatoryes}
Wenxiang Jiao, Wenxuan Wang, Jen tse Huang, Xing Wang, Shuming Shi, and Zhaopeng Tu. 2023.
\newblock \href {http://arxiv.org/abs/2301.08745} {Is chatgpt a good translator? yes with gpt-4 as the engine}.

\bibitem[{Kamoi et~al.(2024)Kamoi, Das, Lou, Ahn, Zhao, Lu, Zhang, Zhang, Zhang, Vummanthala, Dave, Qin, Cohan, Yin, and Zhang}]{kamoi2024evaluatingllmsdetectingerrors}
Ryo Kamoi, Sarkar Snigdha~Sarathi Das, Renze Lou, Jihyun~Janice Ahn, Yilun Zhao, Xiaoxin Lu, Nan Zhang, Yusen Zhang, Ranran~Haoran Zhang, Sujeeth~Reddy Vummanthala, Salika Dave, Shaobo Qin, Arman Cohan, Wenpeng Yin, and Rui Zhang. 2024.
\newblock \href {http://arxiv.org/abs/2404.03602} {Evaluating llms at detecting errors in llm responses}.

\bibitem[{Lommel et~al.(2024)Lommel, Gladkoff, Melby, Wright, Strandvik, Gasova, Vaasa, Benzo, Sparano, Foresi, Innis, Han, and Nenadic}]{lommel2024multirangetheorytranslationquality}
Arle Lommel, Serge Gladkoff, Alan Melby, Sue~Ellen Wright, Ingemar Strandvik, Katerina Gasova, Angelika Vaasa, Andy Benzo, Romina~Marazzato Sparano, Monica Foresi, Johani Innis, Lifeng Han, and Goran Nenadic. 2024.
\newblock \href {http://arxiv.org/abs/2405.16969} {The multi-range theory of translation quality measurement: Mqm scoring models and statistical quality control}.

\bibitem[{Manakhimova et~al.(2023)Manakhimova, Avramidis, Macketanz, Lapshinova-Koltunski, Bagdasarov, and M{\"o}ller}]{manakhimova-etal-2023-linguistically}
Shushen Manakhimova, Eleftherios Avramidis, Vivien Macketanz, Ekaterina Lapshinova-Koltunski, Sergei Bagdasarov, and Sebastian M{\"o}ller. 2023.
\newblock \href {https://doi.org/10.18653/v1/2023.wmt-1.23} {Linguistically motivated evaluation of the 2023 state-of-the-art machine translation: Can {C}hat{GPT} outperform {NMT}?}
\newblock In \emph{Proceedings of the Eighth Conference on Machine Translation}, pages 224--245, Singapore. Association for Computational Linguistics.

\bibitem[{OpenAI(2024)}]{openai2024gpt4o}
OpenAI. 2024.
\newblock \href {https://openai.com/index/gpt-4o-system-card/} {Gpt-4o system card}.
\newblock Accessed: 2025-05-18.

\bibitem[{Papineni et~al.(2002)Papineni, Roukos, Ward, and Zhu}]{papineni-etal-2002-bleu}
Kishore Papineni, Salim Roukos, Todd Ward, and Wei-Jing Zhu. 2002.
\newblock \href {https://doi.org/10.3115/1073083.1073135} {Bleu: a method for automatic evaluation of machine translation}.
\newblock In \emph{Proceedings of the 40th Annual Meeting of the Association for Computational Linguistics}, pages 311--318.

\bibitem[{Peng et~al.(2023)Peng, Ding, Zhong, Shen, Liu, Zhang, Ouyang, and Tao}]{peng2023makingchatgptmachinetranslation}
Keqin Peng, Liang Ding, Qihuang Zhong, Li~Shen, Xuebo Liu, Min Zhang, Yuanxin Ouyang, and Dacheng Tao. 2023.
\newblock \href {http://arxiv.org/abs/2303.13780} {Towards making the most of chatgpt for machine translation}.

\bibitem[{Perrella et~al.(2024)Perrella, Proietti, Cabot, Barba, and Navigli}]{perrella2024correlationinterpretableevaluationmachine}
Stefano Perrella, Lorenzo Proietti, Pere-Lluís~Huguet Cabot, Edoardo Barba, and Roberto Navigli. 2024.
\newblock \href {http://arxiv.org/abs/2410.05183} {Beyond correlation: Interpretable evaluation of machine translation metrics}.

\bibitem[{Qian and Kong(2024)}]{10.1007/978-3-031-60615-1_8}
Ming Qian and Chuiqing Kong. 2024.
\newblock \href {https://doi.org/10.1007/978-3-031-60615-1_8} {Enabling human-centered machine translation using concept-based large language model prompting and translation memory}.
\newblock In \emph{Artificial Intelligence in HCI: 5th International Conference, AI-HCI 2024, Held as Part of the 26th HCI International Conference, HCII 2024, Washington, DC, USA, June 29–July 4, 2024, Proceedings, Part III}, page 118–134, Berlin, Heidelberg. Springer-Verlag.

\bibitem[{Rei et~al.(2020)Rei, Stewart, Farinha, and Lavie}]{rei-etal-2020-comet}
Ricardo Rei, Craig Stewart, Ana~C. Farinha, and Alon Lavie. 2020.
\newblock \href {https://doi.org/10.18653/v1/2020.emnlp-main.213} {Comet: A neural framework for mt evaluation}.
\newblock In \emph{Proceedings of the 2020 Conference on Empirical Methods in Natural Language Processing (EMNLP)}, pages 2685--2702.

\bibitem[{Sellam et~al.(2020)Sellam, Das, and Parikh}]{sellam-etal-2020-bleurt}
Thibault Sellam, Dipanjan Das, and Ankur~P. Parikh. 2020.
\newblock \href {https://doi.org/10.18653/v1/2020.acl-main.704} {Bleurt: Learning robust metrics for text generation}.
\newblock In \emph{Proceedings of the 58th Annual Meeting of the Association for Computational Linguistics}, pages 7881--7892.

\bibitem[{Wu et~al.(2024)Wu, Yuan, Haffari, and Wang}]{wu2024perhapshumantranslationharnessing}
Minghao Wu, Yulin Yuan, Gholamreza Haffari, and Longyue Wang. 2024.
\newblock \href {http://arxiv.org/abs/2405.11804} {(perhaps) beyond human translation: Harnessing multi-agent collaboration for translating ultra-long literary texts}.

\bibitem[{Xu et~al.(2024)Xu, Zhu, Zhao, Pan, Li, and Wang}]{xu2024prideprejudicellmamplifies}
Wenda Xu, Guanglei Zhu, Xuandong Zhao, Liangming Pan, Lei Li, and William~Yang Wang. 2024.
\newblock \href {http://arxiv.org/abs/2402.11436} {Pride and prejudice: Llm amplifies self-bias in self-refinement}.

\bibitem[{Yan et~al.(2024)Yan, Yan, Chen, Li, Zhu, and Zhang}]{yan2024benchmarkinggpt4humantranslators}
Jianhao Yan, Pingchuan Yan, Yulong Chen, Jing Li, Xianchao Zhu, and Yue Zhang. 2024.
\newblock \href {http://arxiv.org/abs/2411.13775} {Benchmarking gpt-4 against human translators: A comprehensive evaluation across languages, domains, and expertise levels}.

\bibitem[{Yan et~al.(2014)Yan, Gao, Pavlick, and Callison-Burch}]{yan-etal-2014-two}
Rui Yan, Mingkun Gao, Ellie Pavlick, and Chris Callison-Burch. 2014.
\newblock \href {https://doi.org/10.3115/v1/P14-1107} {Are two heads better than one? crowdsourced translation via a two-step collaboration of non-professional translators and editors}.
\newblock In \emph{Proceedings of the 52nd Annual Meeting of the Association for Computational Linguistics (Volume 1: Long Papers)}, pages 1134--1144, Baltimore, Maryland. Association for Computational Linguistics.

\end{thebibliography}
\end{document}